\documentclass[11pt]{article}

\usepackage[preprint]{acl}

\usepackage{times}
\usepackage{latexsym}

\usepackage[T1]{fontenc}

\usepackage[utf8]{inputenc}

\usepackage{microtype}

\usepackage{inconsolata}

\usepackage{graphicx}
\usepackage{hyperref}
\usepackage{url}
\usepackage{booktabs}
\usepackage{makecell} 
\usepackage{multirow}
\usepackage{listings}
\usepackage{xcolor} 
\usepackage{upquote}
\usepackage{pifont} 
\usepackage{graphicx}
\usepackage[table]{xcolor}
\usepackage{colortbl}     
\usepackage{graphicx}    
\usepackage{booktabs}
\usepackage{pifont}
\usepackage{amsmath}
\usepackage{paralist}
\usepackage{tabularx}
\usepackage{arydshln}
\usepackage[most]{tcolorbox}
\usepackage{algorithm}
\usepackage{algorithmicx}
\usepackage{algpseudocode}
\usepackage{multirow}
\usepackage{xstring}
\usepackage{longtable}
\usepackage{array} 
\usepackage{booktabs}
\usepackage{float}
\newcommand{\cmark}{\textcolor{green!70!black}{\ding{51}}}%
\newcommand{\xmark}{\textcolor{red}{\ding{55}}}%
\definecolor{myorange}{RGB}{255,165,0}
\definecolor{myblue}{RGB}{0,116,217}
\lstdefinelanguage{json}{
  basicstyle=\ttfamily\footnotesize,
  numbers=left,
  numberstyle=\tiny\color{gray},
  stepnumber=1,
  numbersep=6pt,
  showstringspaces=false,
  keepspaces=true,
  breaklines=true,
  breakatwhitespace=false,
  columns=fullflexible,
  upquote=true,
  postbreak=\mbox{\textcolor{gray}{$\hookrightarrow$}\space},
  literate=
   *{:}{{{\color{red}:}}}{1}
    {,}{{{\color{red},}}}{1}
    {"}{{{\color{orange}"}}}{1}
}

\usepackage{subcaption}
\usepackage{enumitem}

\newtcblisting{jsonbox}{
  enhanced, breakable,
  colback=gray!3,
  colframe=gray!70!black,
  colbacktitle=gray!40,
  coltitle=white,
  coltext=black,
  fonttitle=\bfseries,
  title={Mechanism JSON \hfill \raisebox{0pt}{\Large\textcolor{blue!60}{}}},
  top=2mm, bottom=2mm, left=2mm, right=2mm,
  listing engine=listings,
  listing only,
  listing options={language=json},
  width=\linewidth, boxsep=0pt, sharp corners
}

\newcommand{\perf}[2]{
  \begin{tabular}{@{}c@{}}
    #1 \\
    \IfBeginWith{#2}{+}{%
      {\color{myorange}\scriptsize#2}
    }{
      {\color{myblue}\scriptsize#2}
    }
  \end{tabular}
}

\usepackage{placeins}

\NewDocumentCommand{\heng}
{ mO{} }{\textcolor{red}{\textsuperscript{\textit{Heng}}\textsf{\textbf{\small[#1]}}}}

\NewDocumentCommand{\qingyun}
{ mO{} }{\textcolor{cyan}{\textsuperscript{\textit{Qingyun}}\textsf{\textbf{\small[#1]}}}}

\NewDocumentCommand{\carl}
{ mO{} }{\textcolor{blue}{\textsuperscript{\textit{Carl}}\textsf{\textbf{\small[#1]}}}}

\NewDocumentCommand{\fixed}
{ mO{} }{\textcolor{red}{\textsuperscript{\textit{Fixed}}\textsf{\textbf{\small[#1]}}}}

\title{\includegraphics[width=0.05\linewidth]{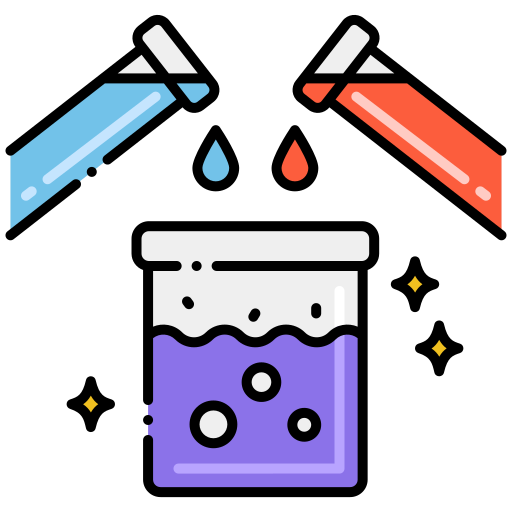} oMeBench: Towards Robust Benchmarking of LLMs in Organic Mechanism Elucidation and Reasoning}

\author{
\textbf{Ruiling Xu\textsuperscript{1}\thanks{\ \ These authors contributed equally to this work.}} \quad
\textbf{Yifan Zhang\textsuperscript{1}\footnotemark[1]}\\
\textsuperscript{1}University of Illinois Urbana-Champaign \quad
\texttt{\{yifanz28, ruiling3\}@illinois.edu} \\
}

\begin{document}
\maketitle
\maketitle

\begin{abstract}
    Organic reaction mechanisms describe the stepwise elementary processes by which reactants transform into intermediates and products, and are fundamental to understanding chemical reactivity and guiding molecular and reaction design. While large language models (LLMs) have shown promise on chemical tasks such as synthesis design, it remains unclear to what extent this reflects genuine chemical reasoning capabilities: the ability to generate chemically valid intermediates, maintain consistency across reaction steps, and follow logically coherent multi-step pathways. To investigate this, we introduce \textbf{oMeBench}, the first large-scale, expert-curated benchmark for organic mechanism reasoning, comprising over 10,000 annotated mechanistic steps with reaction type labels, intermediate structures, and difficulty ratings. To enable fine-grained evaluation, we further propose \textbf{oMeS}, a dynamic scoring framework that jointly assesses step-level logical consistency and chemical structural similarity. Systematic evaluation of state-of-the-art LLMs reveals that while current models exhibit promising chemical intuition, they struggle to produce correct and consistent reasoning across multi-step mechanisms. Notably, combining prompting strategies with fine-tuning enables smaller-scale models to achieve performance comparable to closed-source frontier models. We hope \textbf{oMeBench} will serve as a rigorous foundation for advancing AI systems toward genuine chemical reasoning.\footnote{https://anonymous.4open.science/r/oMeBench-E28D}
\end{abstract}

\section{Introduction}
\label{sec:introduction}
\begin{figure}[t]
    \centering
    \includegraphics[width=\linewidth]{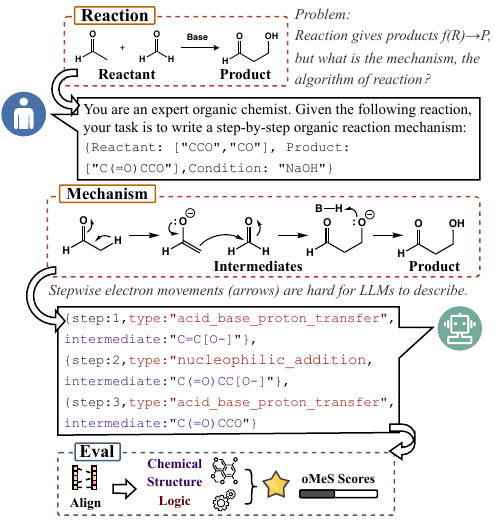}
    \caption{Given a reaction from oMeBench, we generate a reaction mechanism and score it with oMeS.}
    \label{fig:teasing}
    \vspace{-5mm}
\end{figure}


\textit{Organic reaction mechanisms}, the stepwise elementary reactions where reactants form intermediates and products, provide a detailed, step-by-step description of how chemical bonds are broken and formed during an organic chemical reaction~\citep{muller1994glossary}. In other words, a mechanism serves as the natural algorithm of a reaction. Mechanisms form the foundation for understanding chemical reactions~\citep{grossman2003mechanisms}, and play a key role in accelerating molecular discovery~\citep{carey2007advanced, smith2025march}. For instance, in drug discovery, mechanistic models serve as the basis for predicting metabolic pathways and toxicities~\citep{silverman2014organic,patrick2023introduction}. Thus, organic reaction mechanisms are not merely theories of fundamental chemistry but tools that drive innovations in applied chemistry.

\begin{figure*}[!htb]
\centering
  \includegraphics[width=0.86\linewidth]{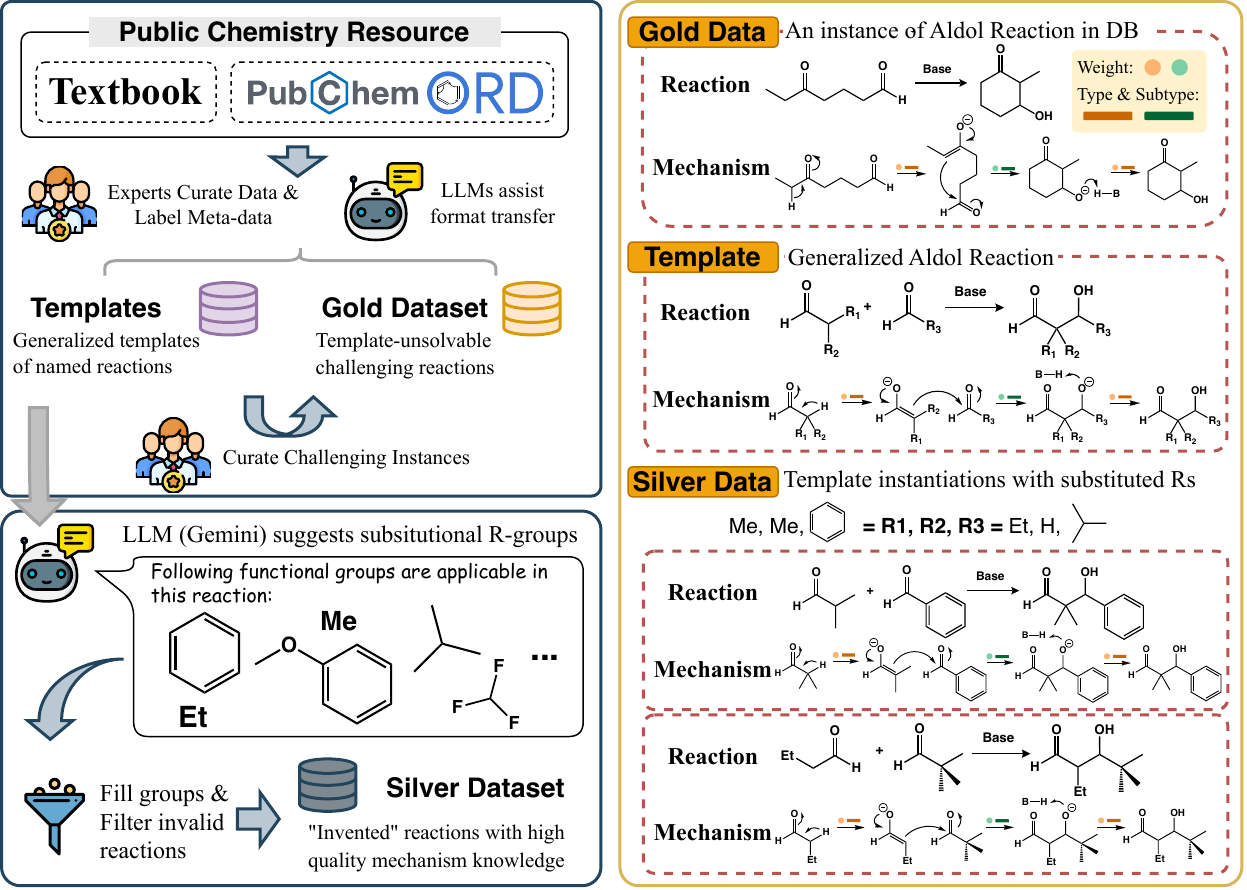}
  \vspace{-4pt}
  \caption{Overview of dataset construction. 
  Each named reaction represents a family of reactions that follow the same mechanistic logic, making them a natural fit for templating. The templates use R-group placeholders to represent variable substituents (e.g., Me, Et, H).
}
  \label{fig:pipeline_data}
  \vspace{-5mm}
\end{figure*}
Large language models (LLMs) have recently demonstrated remarkable performance in diverse chemical applications, such as property prediction~\citep{ahmad2022chemberta, stark2022equibindgeometricdeeplearning}, retrosynthetic analysis~\citep{chen2023g}, product prediction~\citep{schwaller2021mapping,jablonka2024leveraging}, reaction mining~\cite{reactionminer2023}, action extraction~\cite{actionie2024}, and reaction-condition optimization~\citep{boiko2023autonomous}.  
Despite this progress, chemical reasoning, especially \textit{mechanism reasoning}, remains largely unexplored. 
Most existing benchmarks stop at mapping reactants to products~\citep{lowe2012extraction, kearnes2021open}, with little to no annotation of the underlying elementary steps. Therefore, models can often predict what happens, but struggle to explain the mechanistic logic behind it. 
Moreover, prior attempts to construct reaction mechanism datasets largely rely on rule-based systems~\citep{chen2024large}, which encode a limited set of hard-coded transformation rules (e.g., deprotection). This results in relatively simple datasets with low reaction diversity. Other benchmarks with natural language reasoning like SciBench~\citep{wang2024scibenchevaluatingcollegelevelscientific} and ChemCoTBench~\citep{li2025beyond} lack explicit molecule-level details, limiting their ability to test mechanism elucidation.

To address this gap, we introduce \textbf{oMeBench} (Fig.~\ref{fig:teasing}), which contains over 10,000 annotated mechanistic steps with intermediates, step-type labels, and rationales, as well as a wide range of reaction classes and difficulty levels. 
For fine-grained evaluation, we further propose \textbf{oMeS}, a dynamic scoring framework that matches predicted mechanisms against gold-standard ones using weighted similarity metrics. Rather than simply checking whether the final product is correct, this lets us look more closely at how a model actually reasons through a reaction, and quantify how mechanistically faithful its predictions are  (mechanism-level accuracy).
Our extensive evaluation reveals that while LLMs demonstrate non-trivial chemical intuition, they struggle to sustain multi-step causal logic, especially in complex or lengthy mechanisms. Importantly, exemplar-based in-context learning and supervised fine-tuning both yield substantial improvements~\citep{dong2022survey, zhang2025instructiontuninglargelanguage}, highlighting not only the potential but also the current limitations of LLMs in mechanistic chemistry. We hope that oMeBench will serve as a rigorous foundation for advancing AI systems toward mechanistic chemical reasoning. 
Our contributions are threefold:
\vspace{-1.5mm}
\begin{itemize}[leftmargin=*, noitemsep]
    \item We construct the first expert-annotated large-scale organic reaction mechanism datasets, with step-by-step mechanisms and natural language rationales, all released in a standardized format that is readable for both human and machine.
    \item We introduce novel dynamic and chemically interpretable evaluation metrics that automatically measure LLMs’ mechanism elucidation and reasoning.
    \item We conduct a systematic study of mechanistic reasoning across multiple LLMs, revealing critical gaps and partially bridging them, achieving performance comparable with proprietary closed-source models.
\end{itemize}

\begin{table*}[!htb]
\centering
\small
\begin{tabularx}{\textwidth}{
  >{\arraybackslash\hsize=2.8\hsize}X
  >{\arraybackslash\hsize=0.5\hsize}X
  >{\arraybackslash\hsize=0.9\hsize}X
  >{\arraybackslash\hsize=1.5\hsize}X
  >{\arraybackslash\hsize=0.8\hsize}X
  >{\arraybackslash\hsize=0.5\hsize}X
  >{\arraybackslash\hsize=0.3\hsize}X
  >{\arraybackslash\hsize=0.3\hsize}X
}
\toprule
\textbf{Dataset} & \textbf{\#Temps} & \textbf{Size(Scaled)} & \textbf{Scaling Method} & \textbf{Level} & \textbf{Mech?} & \textbf{Int?} & \textbf{NL?} \\
\midrule
USPTO-50k\citep{lowe2012extraction} & 86  & 50k   & Expert System   & Easy   & \xmark & \xmark & \xmark \\
\citet{chen2024large}       & 86  & 31k   & Expert System   & Easy   & \cmark & \xmark & \xmark \\
\citet{joung2024reproducing} & 86  & 19.2k & Expert System   & Easy   & \cmark & \cmark & \xmark \\
RMechDB\citep{tavakoli2023rmechdb} & --  & 5k+   & --              & Medium & \cmark & \cmark & \xmark \\
ChemCoTBench\citep{li2025beyond}        & 100 & 100   & Manual Curation & Easy   & \cmark & \xmark & \cmark \\
SciBench\citep{wang2024scibenchevaluatingcollegelevelscientific} & 30  & 30    & Manual Curation & Easy   & \xmark & \xmark & \cmark \\
\textbf{Ours}               & 196 & 10k   & Manual Curation & Mixed  & \cmark & \cmark & \cmark \\
\bottomrule
\end{tabularx}
\vspace{-1mm}
\caption{Comparison of representative organic reaction datasets by scale, construction method, and annotation type. "\#Temps" denotes unique templates; "Mech?", "Int?", and "NL?" indicate the presence of mechanistic details, intermediates, and natural language descriptions, respectively.} 
\label{tab:dataset_comparison}
\vspace{-1mm}
\end{table*}

\begin{table*}[htbp]
\centering
\small
\begin{tabularx}{\textwidth}{
  >{\arraybackslash\hsize=0.9\hsize}X
  >{\arraybackslash\hsize=0.6\hsize}X
  >{\arraybackslash\hsize=0.4\hsize}X
  >{\arraybackslash\hsize=0.3\hsize}X
  >{\arraybackslash\hsize=0.5\hsize}X
  >{\arraybackslash\hsize=3.3\hsize}X
}
\toprule
\textbf{Dataset} & \textbf{\#Reactions} & \textbf{\#Steps} & \textbf{\#Types} & \textbf{\#Subtypes} & \textbf{Annotation Source} \\ 
\midrule
\textbf{oMe-Gold}     & 196   & 858    & 8   & 30  & Expert-verified, curated from textbooks and literature \\
\textbf{oMe-Template} & 167   & 722    & 8   & 30  & Expert-curated templates abstracted from oMe-Gold \\
\textbf{oMe-Silver}   & 2,508 & 10,619 & 8   & 30  & Automatically expanded from oMe-Template with filtering \\
\bottomrule
\end{tabularx}
\vspace{-3mm}
\caption{Statistics of the three curated datasets in oMeBench.}
\label{tab:dataset_statistics}
\vspace{-2mm}
\end{table*}


\section{Related Work}
\vspace{-2mm}
\label{sec:related_works}

\noindent\textbf{Organic Reaction Datasets.}
Large-scale reaction datasets, such as USPTO \citep{lowe2012extraction}, Pistachio \citep{mayfield2018pistachio}, and PubChem \citep{kim2025pubchem}, collect reactions from the literature without providing explicit mechanistic details. Recent mechanism-aware datasets generally rely on atom-mapped encodings that are difficult to interpret (App.~\ref{app:data_format_comparion}), and remain limited in scope: some focus narrowly on radical \citep{tavakoli2023rmechdb} or polar reactions \citep{tavakoli2024pmechdb}, while others are oversimplified with low diversity \citep{chen2024large, joung2024reproducing}. Broader scientific benchmarks \citep{li2025beyond,wang2024scibenchevaluatingcollegelevelscientific} include mechanisms only in simple QA formats.

\noindent\textbf{Chemistry-Specific Models.}
Domain-specific models, such as ChemDFM~\citep{zhao2024chemdfm}, leverage chemical corpora through pretraining and fine-tuning, while Ether0~\citep{narayanan2025trainingscientificreasoningmodel} improves multi-step reasoning with supervised and reinforcement learning. 
mCLM \cite{edwards2025mclmfunctioninfusedsynthesisfriendlymodular} pushes this further by working at the representation level, treating molecules as modular functional building blocks rather than monolithic structures. Despite these advances, most of these models are still oriented toward tasks like molecular property prediction, retrosynthesis, or text-based QA. Explicit mechanistic reasoning remains largely unaddressed.


\noindent\textbf{Chemistry-related tasks in LLMs.}
Research on LLMs in chemistry has evolved from early tasks such as molecular name translation and property prediction \citep{chithrananda2020chemberta,zheng2024llm4sd} to more complex reasoning applications, including retrosynthesis and automated synthesis protocols \citep{wang2025llm,cadeddu2024language}. Recent work has explored prompting strategies and autonomous agents to extend model capabilities \citep{bran2024chemcrow,darvish2025organa}. However, LLM performance on these tasks remains limited. In particular, understanding organic mechanisms, the foundation of all organic transformations, has received little attention, with prior papers addressing only a few simple cases \citep{bran2025chemicalreasoningllmsunlocks}. To address this gap, we present the first comprehensive study of LLMs on mechanism reasoning and release a large-scale dataset to support fine-tuning and evaluation.

\section{Dataset}
\vspace{-1mm}
\label{sec:dataset}

\begin{figure*}[!htb]
  \centering
  \subfloat{
    \includegraphics[width=0.98\linewidth]{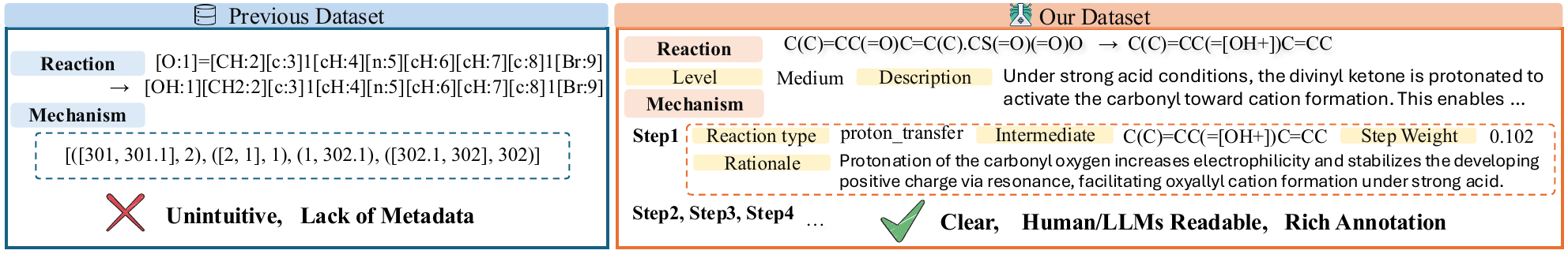}
    \label{fig:data_format}
  }\\
  \setcounter{subfigure}{3}
{
    \includegraphics[width=0.95\linewidth]{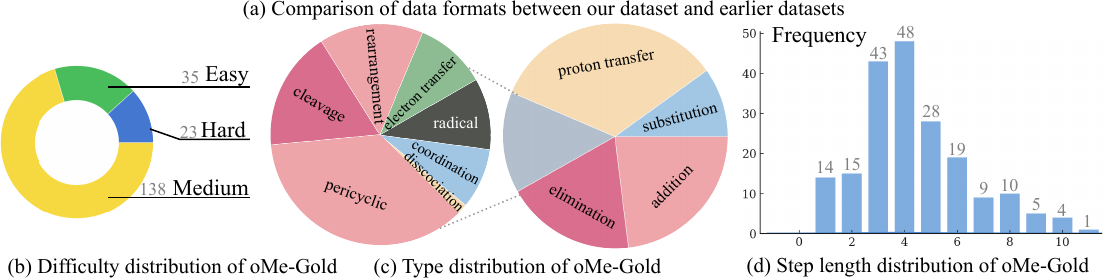}
    \label{fig:pipeline_data_gold}
    \vspace{-.5mm}
  }
  \vspace{-2mm}
  \caption{Overview of the dataset metadata and format.}
  \vspace{-5mm}
\end{figure*}

Tab.~\ref{tab:dataset_comparison} highlights three key limitations of early datasets:
(i) \textbf{unreadable formats}. Each mechanistic step is represented through atom-mapping, which is non-intuitive and difficult to interpret (App.~\ref{app:data_format_comparion}); (ii) \textbf{missing metadata}, limiting deeper analysis; and
(iii) \textbf{lack of difficulty and diversity}, as most reactions are either high-school level, domain-specific, or repetitive despite large scale.

To address these limitations, we introduce three complementary organic reaction mechanism datasets.
\textbf{oMe-Gold} comprises literature-verified organic reactions with detailed mechanisms, serving as the gold-standard benchmark for evaluation. \textbf{oMe-Template} consists of mechanistic templates that generalize families of reactions sharing a common mechanistic pathway. Each template contains substitutable placeholder \textit{R-groups},
 enabling instantiation into diverse reaction instances. \textbf{oMe-Silver} is a large-scale dataset systematically expanded from \textbf{oMe-Template}, containing chemically plausible but not necessarily literature-verified reactions, and primarily used for large-scale model training. All three datasets include rich expert annotations covering difficulty. Tab.~\ref{tab:dataset_statistics} summarizes the dataset statistics, and Fig.~\ref{fig:pipeline_data} illustrates the complete construction process.

\subsection{Gold Dataset}
All reactions were curated from authoritative organic chemistry textbooks and reaction databases \citep{kim2025pubchem,kearnes2021open,grossman2003mechanisms,kurti2005strategic}. For each named reaction\footnote{A named reaction refers to a class of reactions following a shared mechanistic template.}, we included either the classical instance or a challenging variant. 
In addition, we included several out-of-domain reactions from textbooks to improve diversity. Reactions were initially extracted using Gemini-Pro-2.5~\citep{team2023gemini} and subsequently verified by experts for SMILES validity and chemical consistency\footnote{Among the 196 entries, 189 required manual correction. Manual verification was essential, as SOTA LLMs and other tools often fail to produce chemically valid results.}.

\noindent\textbf{Difficulty Level.}
We define difficulty based on mechanistic complexity and the reasoning depth. 
\textbf{Easy} reactions make up about 20\% of the dataset and follow straightforward, single-step logic. Most reactions (70\%) sit in the \textbf{Medium} tier, where some conditional reasoning is needed. The \textbf{Hard} 10\% involve genuinely complex mechanisms, which would challenge even an experienced chemist.
All reactions were independently annotated by two expert chemists, with disagreements resolved by a third expert (App.~\ref{app:annotator}).

\noindent\textbf{Step Type and Subtype.}
We propose a two-level taxonomy for elementary steps, comprising 8 types and 30 subtypes, detailed in Tab.~\ref{tab:types_subtypes_grouped}. The annotation procedure follows \textit{Difficulty Level}.

\noindent\textbf{Descriptions and Rationales.}
Each entry includes a high-level natural language description of the reaction and a rationale for every mechanistic step explaining the underlying mechanistic principles. These texts were generated using Gemini-Pro-2.5 through prompts in App.~\ref{app:prompt_rationale}.

\noindent\textbf{Step Weighting.} 
Each step is weighted by its mechanistic importance and reasoning complexity. We 1) assign initial weights to each step type, 2) adjust for unusually simple or difficult subtypes, and then 3) fine-tune by positional or structural factors. The complete details are in App.~\ref{app:weighting_strategy}.

\begin{figure*}[!htb]
  \includegraphics[width=1.0\linewidth]{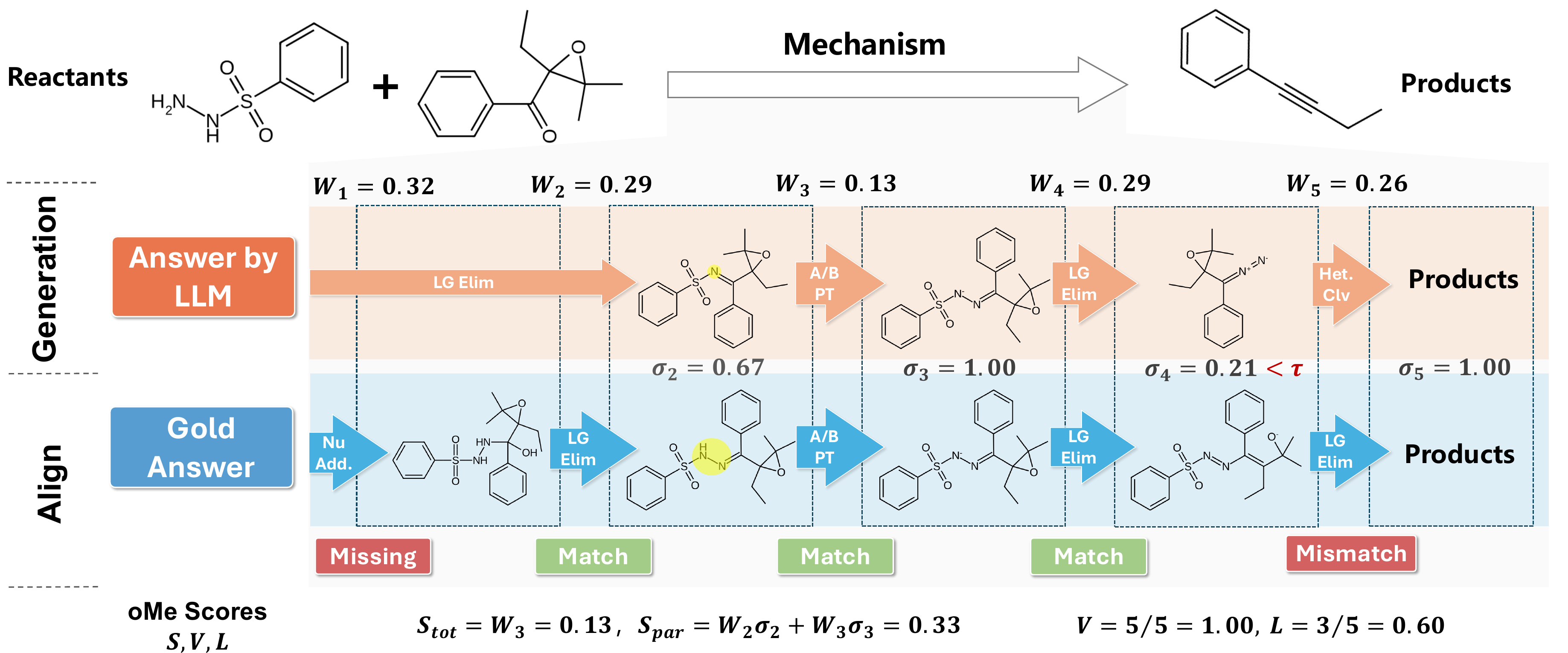}
  \vspace{-3mm}
  \caption {The overview of the evaluation system. LLM-generated mechanisms are dynamically aligned with gold references. Subtype correctness, molecular similarity $\sigma$, and weight $W$ are used to compute oMeS $S$, $V$, and $L$.}
  \label{fig:pipeline}
  \vspace{-5mm}
\end{figure*}
\subsection{Template Abstraction and Expansion}
\noindent\textbf{oMe-Template} comprises template-level reactions from named reaction databases, which provide generic \textit{R-group} structures and representative instances. As these sources lack explicit mechanistic details, expert chemists reconstructed each mechanism by generalizing from the instance-level reaction while preserving its reaction conditions. This abstraction is chemically justified, since named reactions inherently represent families of reactions sharing common mechanistic pathways. Metadata annotation followed the same protocol as \textbf{oMe-Gold}. Representative examples are in App.~\ref{app:data_format_comparion}.

\noindent\textbf{oMe-Silver} is constructed by expanding \textbf{oMe-Template}. For each template, Gemini-Pro-2.5 proposes chemically plausible R-group substitutions (App.~\ref{app:prompt_suggest_groups}), and invalid SMILES are filtered automatically. Metadata are inherited from the source template, and the scaling algorithm is in App.~\ref{app:scaling_algorithm}. 
On average, each template produces 15 valid silver-standard reactions. Compared to simple rule-based substitution, this LLM-guided expansion tends to capture a wider range of chemically diverse and plausible structures.
To prevent data contamination between the training and evaluation sets, we filter out exact matches during construction. A nearest-neighbor analysis further shows that Silver reactions do not duplicate or closely shadow any Gold reactions (App.~\ref{app:contamination}).

\section{Evaluation}
\vspace{-2mm}
\label{Evaluation}

We introduce \textbf{oMeS}, a dynamic evaluation framework with four complementary metrics for fine-grained assessment of LLMs' mechanistic reasoning. It addresses three main challenges: (i) unreliable LLM-as-judge results due to weak domain knowledge, (ii) the limitation of static string matching for complex mechanistic outputs, and (iii) the need to score partial reasoning. \textbf{oMeS} resolves this through dynamic alignment (\S~\ref{sec:alignment}) and weighted partial scoring (\S~\ref{sec:definition}).
The evaluation pipeline (Fig.~\ref{fig:pipeline}) consists of three stages:
(i) Generate: Prompt the LLM to produce stepwise mechanisms, each with an intermediate (\textsc{SMILES}) and selected reaction subtype (Tab.~\ref{tab:types_subtypes_grouped}). The full prompt is in App.~\ref{app:prompt_default}. (ii) Align: Match the predicted and gold mechanisms using a weighted Needleman–Wunsch dynamic algorithm~\citep{likic2008needleman}. (iii) Calculate: Derive four complementary metrics from the alignment to enable detailed, fine-grained evaluation of mechanistic reasoning.

\subsection{Definitions}
\label{sec:definition}
Let the gold mechanism be $G=\{(t_k, I_k, w_k)\}_{k=1}^{N}$ with $N$ steps, where $t_k$ is the gold subtype label, $I_k$ is the gold intermediate \textsc{SMILES}, and $w_k\!\ge\!0$ is the step weight with $\sum_{k=1}^{N} w_k = 1$. Let the model's raw prediction be $P=\{(\hat t_j,\hat I_j)\}_{j=1}^{M}$ with $M$ predicted intermediates. After alignment (\S~\ref{sec:alignment}), each gold step $k$ is matched to at most one predicted step; we denote such an aligned prediction at position $k$ as $(\hat t_k,\hat I_k)$ or marked missing if unmatched.

\noindent\textbf{SMILES Validity Score ($V$)} measures a model's ability to produce chemically valid intermediates, defined as the fraction of predicted \textsc{SMILES} that can be parsed by RDKit.

\noindent\textbf{Logical Fidelity Score ($L$)} measures stepwise mechanistic labeling accuracy irrespective of structural correctness, computed over aligned positions. Formal definitions of $V$ and $L$ are in App.~\ref{app:metrics_details}.

\noindent\textbf{Organic Mechanism Scores ($S_{tot}, S_{part}$)} measures LLMs' reasoning ability in organic reaction mechanisms. For each step, subtype correctness is a prerequisite, since 
the model can demonstrate understanding of the underlying chemical rationale only with the correct subtype.
\textbf{oMeS-total} ($S_{\text{tot}}$) will assign credit only for \emph{exact} matches:
\vspace{-2mm}
\[
s_k^{\text{tot}} = 
w_k \, \mathbf{1}_{\{t_k = \hat t_k\}} \, \mathbf{1}_{\{I_k = \hat I_k\}},
\qquad
S_{\text{tot}} = \sum_{k=1}^{N} s_k^{\text{tot}}
\vspace{-1mm}
\]

\begin{table*}[!htb]
\centering

\small
\resizebox{\textwidth}{!}{%
\begin{tabular}{@{}l r rrrr rrrr rrrr rrrr@{}}
\toprule
\multirow{2}{*}{\textbf{Model}} & \multirow{2}{*}{\textbf{Size}} & \multicolumn{4}{c}{\textbf{Overall}} & \multicolumn{4}{c}{\textbf{Easy}} & \multicolumn{4}{c}{\textbf{Medium}} & \multicolumn{4}{c}{\textbf{Hard}} \\
\cmidrule(lr){3-6} \cmidrule(lr){7-10} \cmidrule(lr){11-14} \cmidrule(lr){15-18}
 & & \textbf{V} & \textbf{L} & \textbf{$S_{tot}$} & \textbf{$S_{part}$} & \textbf{V} & \textbf{L} & \textbf{$S_{tot}$} & \textbf{$S_{part}$} & \textbf{V} & \textbf{L} & \textbf{$S_{tot}$} & \textbf{$S_{part}$} & \textbf{V} & \textbf{L} & \textbf{$S_{tot}$} & \textbf{$S_{part}$} \\
\midrule
\multicolumn{18}{c}{\textbf{Proprietary and Frontier Models}} \\
\midrule
GPT-5.5 & - & \textbf{99.2} & \textbf{66.6} & \underline{42.6} & \underline{48.2} & \textbf{100.0} & \textbf{86.7} & 59.6 & \underline{71.6} & \underline{98.8} & \textbf{64.3} & \underline{41.2} & \underline{45.2} & \textbf{100.0} & \textbf{50.0} & \underline{25.3} & \underline{30.8} \\
Claude-Opus-4.7 & - & 94.2 & 53.1 & 29.1 & 33.6 & 96.1 & 67.1 & 37.7 & 46.0 & 94.5 & 53.4 & 30.0 & 33.8 & 89.3 & 30.4 & 11.1 & 13.9 \\
Claude-Sonnet-4.6 & - & 91.8 & 51.5 & 25.4 & 30.5 & 95.9 & 62.5 & 39.4 & 43.4 & 90.9 & 51.9 & 23.8 & 29.8 & 90.9 & 32.9 & 13.3 & 15.1 \\
Gemini-3.1-Pro-Preview & - & \textbf{99.2} & \underline{63.5} & \textbf{47.6} & \textbf{51.3} & \textbf{100.0} & \underline{83.8} & \textbf{71.5} & \textbf{77.1} & \textbf{99.3} & \underline{60.8} & \textbf{44.1} & \textbf{47.1} & \underline{97.0} & \underline{49.1} & \textbf{32.6} & \textbf{36.9} \\
Gemini-2.5-Pro    & -   & 90.5 & 57.8 & 34.9 & 37.9 & 94.5 & 70.5 & 56.5 & 61.3 & 90.3 & 56.6 & 32.4 & 35.2 & 85.4 & 46.3 & 16.6 & 18.5 \\
DeepSeek-V4-Pro & - & \underline{98.0} & 61.6 & 41.4 & 45.2 & \textbf{100.0} & 79.5 & \underline{60.7} & 65.8 & 97.9 & 59.2 & 39.6 & 42.8 & 95.3 & 48.3 & 22.7 & 28.2 \\
DeepSeek-V4-Flash & - & 95.6 & 51.3 & 30.8 & 34.0 & \underline{96.9} & 67.3 & 46.9 & 48.5 & 96.1 & 51.4 & 30.4 & 34.1 & 90.2 & 26.2 & 9.0 & 11.8 \\
\midrule
\multicolumn{18}{c}{\textbf{Open Source Models}} \\
\midrule
GPT-OSS & 20B & 78.2 & 31.5 & 12.9 & 14.2 & 88.4 & 47.6 & 25.1 & 28.0 & 77.0 & 30.3 & 11.7 & 12.6 & 70.2 & 14.8 & 1.8 & 2.4 \\
LLaMA-3 & 8B & 73.5 & 12.6 & 2.7 & 3.3 & 69.9 & 13.0 & 1.5 & 1.5 & 74.2 & 13.2 & 3.2 & 4.0 & 74.9 & 7.9 & 1.5 & 1.8 \\
Qwen-3 & 4B & 58.4 & 14.9 & 3.2 & 4.2 & 63.1 & 12.1 & 2.6 & 2.6 & 60.4 & 17.5 & 3.8 & 5.3 & 40.3 & 4.3 & 0.2 & 0.3 \\
Mistral & 7B & 52.8 & 12.7 & 2.0 & 2.2 & 56.6 & 14.6 & 1.7 &  2.2 & 52.5 & 12.8 & 2.3 & 2.5 & 48.6 & 9.0 & 1.0 & 1.0 \\
Phi-4 & 7B & 0.3 & 0.0 & 0.0 & 0.0 & 0.0 & 0.0 & 0.0 & 0.0 & 0.3 & 0.0 & 0.0 & 0.0 & 0.0 & 0.0 & 0.0 & 0.0 \\
\midrule
\multicolumn{18}{c}{\textbf{Domain-Specific Models}} \\
\midrule
OLMo-2 & 13B & 17.4 & 6.1 & 0.6 & 0.6 & 22.0 & 10.2 & 3.1 & 3.1 & 16.5 & 5.2 & 0.0 & 0.0 & 16.2 & 4.9 & 0.0 & 0.0 \\
OpenBioLLM & 8B & 4.4 & 1.7 & 0.3 & 0.3 & 9.4 & 1.7 & 0.3 & 0.3 & 3.8 & 1.9 & 0.3 & 0.3 & 0.0 & 0.0 & 0.0 & 0.0 \\
ChemDFM & 8B & 0.0 & 0.7 & 0.0 & 0.0 & 0.0 & 0.0 & 0.0 & 0.0 & 0.7 & 0.0 & 0.0 & 0.0 & 0.0 & 0.0 & 0.0 & 0.0 \\
\bottomrule
\end{tabular}
}
\caption{Model Performance on oMeBench Across Difficulty Levels}
\label{tab:results}
\vspace{-5mm}
\end{table*}

\noindent\textbf{oMeS-partial} ($S_{\text{part}}$) grants partial credit when the generated intermediate has meaningful chemical similarity to the gold intermediate, as measured by molecular fingerprints, while \textbf{oMeS-total} is stricter. 
Let $\mathbf{T}$ denotes the \textit{Tanimoto similarity}~\citep{bajusz2015tanimoto} of \textit{Morgan fingerprints}~\citep{rogers2010extended} and $\tau$ is a similarity threshold:
\vspace{-1mm}
\[
\sigma(I_k,\hat I_k) = \mathbf{T}(I_k,\hat I_k) \cdot \mathbf{1}_{\{\mathbf{T}(I_k,\hat I_k) \ge \tau\}}
\]
\vspace{-4mm}
\[
S_{\text{part}} = \sum_{k=1}^{N} s_k^{\text{par}}; \quad s_k^{\text{par}} = w_k \, \sigma(I_k,\hat I_k) \, \mathbf{1}_{\{t_k = \hat t_k\}}
\]
We find that \textbf{oMeS} is robust to the choice of similarity metric. Rankings remain nearly identical when using graph-based similarity instead (App.~\ref{app:similarity_ablation}). We adopt Tanimoto on Morgan fingerprints because it directly reflects local bond changes, which are central to mechanistic reasoning.

\vspace{-1mm}
\subsection{Alignment Strategy}
\label{sec:alignment}
We align $G=\{(t_k, I_k, w_k)\}_{k=1}^{N}$ and $P=\{(\hat t_j,\hat I_j)\}_{j=1}^{M}$ with Needleman--Wunsch algorithm:
\begin{compactitem}

    \item \textbf{Match:} if $t_k=\hat t_j$, we evaluate both exact-match ($s^{\text{tot}}$) and similarity-based credit ($s^{\text{par}}$);
    \item \textbf{Type mismatch:} if $t_k\neq \hat t_j$, we assign zero credit and a mild penalty;
    \item \textbf{Gaps (skip):} skipping a gold step or a predicted step incurs a fixed penalty.
\end{compactitem}

To break ties, we prioritize alignment paths sequentially by (1) higher $S_{\text{tot}}$, (2) higher $S_{\text{part}}$, (3) fewer mismatches, and (4) fewer gaps. The final traceback assigns one of four step-level tags: \texttt{match}, \texttt{type\_mismatch}, \texttt{skip\_gold} (missing prediction), or \texttt{extra} (spurious prediction).
The alignment procedure is robust to common edge cases, including minor \textsc{SMILES} typos or benign variants, skipped or extra steps, and cases where the intermediate is plausible but the subtype is mislabeled. Case studies and ablations are in App.~\ref{app:alignment}.

\section{Experiments}
\label{sec:experiments}
\subsection{Experimental Setup.} To establish the baseline performance on our benchmark, we evaluate multiple different LLMs under a common one-shot prompting setup. We adopt four metrics: \textbf{Validity (V)}, \textbf{Logicality (L)}, \textbf{oMeS-total ($S_{tot}$)} and \textbf{oMeS-partial ($S_{part}$)}. Detailed prompts are in App.~\ref{app:prompt_default}. The results are presented in Tab.~\ref{tab:results} and Fig.~\ref{fig:heatmap}.

\noindent\textbf{Evaluated Models.} We categorize LLMs into three groups: (i) \textbf{Proprietary LLMs}: GPT-5.5~\citep{singh2025openai}, Claude series~\citep{anthropic2026claudeopus, anthropic2026claudesonnet}, and Gemini 2.5 Pro and 3.1 Pro Preview~\citep{team2023gemini}; (ii) \textbf{Open-source LLMs}: DeepSeek-V4 series~\citep{deepseekai2026deepseekv4} and instruct version of LLaMa-3~\citep{grattafiori2024llama}, Qwen-3~\citep{yang2025qwen3}, Mistral, and Phi-4-mini~\citep{abdin2024phi}; (iii) \textbf{Domain-specific LLMs}: OLMo-2~\citep{olmo20242}, OpenBioLLM~\citep{OpenBioLLMs}, ChemDFM~\citep{zhao2024chemdfm}.

\subsection{LLMs show limited understanding of mechanisms} 
The evaluation indicates that while current LLMs often produce syntactically valid SMILES and loosely follow sequences of reaction types, their understanding of reaction mechanisms remains highly limited. Even the strongest frontier models achieve only moderate scores that decline sharply with task difficulty. The best-performing model, \textit{Gemini-3.1-Pro-Preview}, reaches only 51.3 on $S_{part}$, while many other models yield scores close to zero, reflecting a profound lack of organic mechanism knowledge. LLMs can imitate the surface form of mechanism elucidation but fail to capture multi-step causal logic underlying organic reactions.

\noindent\textbf{Domain-specific models underperform.}  
Although chemistry-trained models (e.g., ChemDFM, OLMo-2) are optimized on large chemical corpora, their performance lags behind that of general-purpose LLMs. They even underperform all open-source baselines across all metrics, even validity. This disparity suggests that while domain pretraining includes vast chemical literature, it fails to provide explicit mechanistic reasoning. Frequent generation of irrelevant or inconsistent structure indicates that these models capture surface-level patterns rather than causal logic in mechanism.

\noindent\textbf{Chain-of-Thought fails to enhance reasoning.} 
One-shot Chain-of-Thought (CoT) prompting~\citep{wei2022chain}  did not yield measurable improvements (Tab.~\ref{tab:oneshot_cot_results}). The generated reasoning are verbose but shallow, merely paraphrasing reaction conditions without offering causal insights. Unlike symbolic or mathematical reasoning, mechanism elucidation requires a structural understanding, which CoT, without chemistry-aware supervision, fails to provide.

\vspace{-1mm}
\subsection{Why do LLMs fail?}

\begin{figure}[hb]
  \centering
  \vspace{-3mm}
  \includegraphics[width=0.87\linewidth]{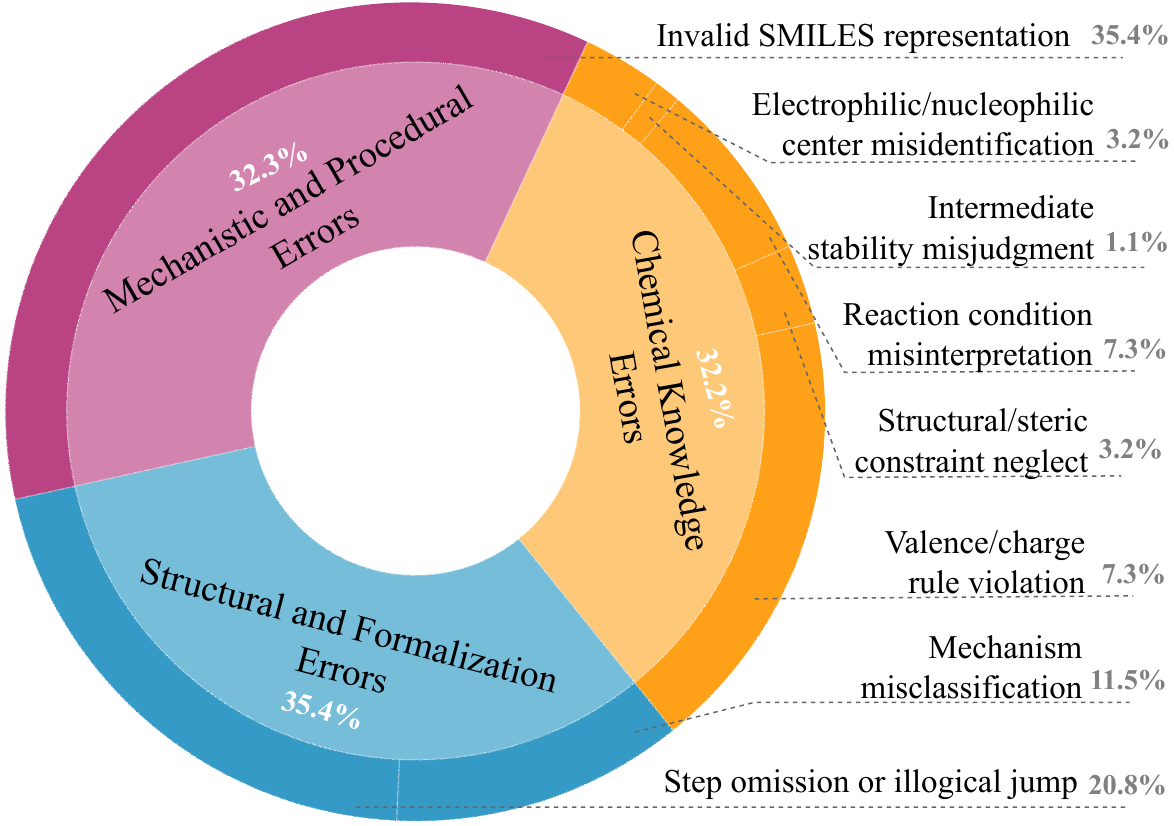}
  \vspace{-3mm}
  \caption{Distribution of mechanistic reasoning errors for Gemini-2.5-Pro. The chart groups errors into three primary categories, with each bar indicating the percentage of a specific subtype.}
  \label{fig:error_analysis}
  \vspace{-2mm}
\end{figure}

To systematically analyze the failure modes, we categorize three broad categories of errors: \emph{Chemical Knowledge Errors}, \emph{Mechanistic and Procedural Errors}, and \emph{Structural and Formalization Errors}. Each of them is further divided into representative subtypes (Tab.~\ref{tab:error_taxonomy_app}). 
These categories are not independent.
For instance, misinterpreting reaction conditions can lead to misidentified reactive centers or invalid intermediates. In Fig.~\ref{fig:error_analysis}, we analyze Gemini-2.5-pro as a representative.

\noindent\textbf{Chemical structural inconsistency.}  
Despite high product-level SMILES validity in Tab.~\ref{tab:results}, 35.4\% of all errors arise from invalid or inconsistent SMILES within multi-step mechanisms. Models can often produce valid final products but fail to maintain consistent intermediate structure across steps, such as unchanged substructure and the number of electrons, where valence and charge violations account for 7.3\% of total errors. The case in Tab.~\ref{tab:case2} shows how LLMs invent inconsistent chemical structures to provide a logical mechanism, which is considered as a \textbf{perceptual gap} of LLMs.

\begin{figure}[ht]
    \centering
    \includegraphics[width=\linewidth]{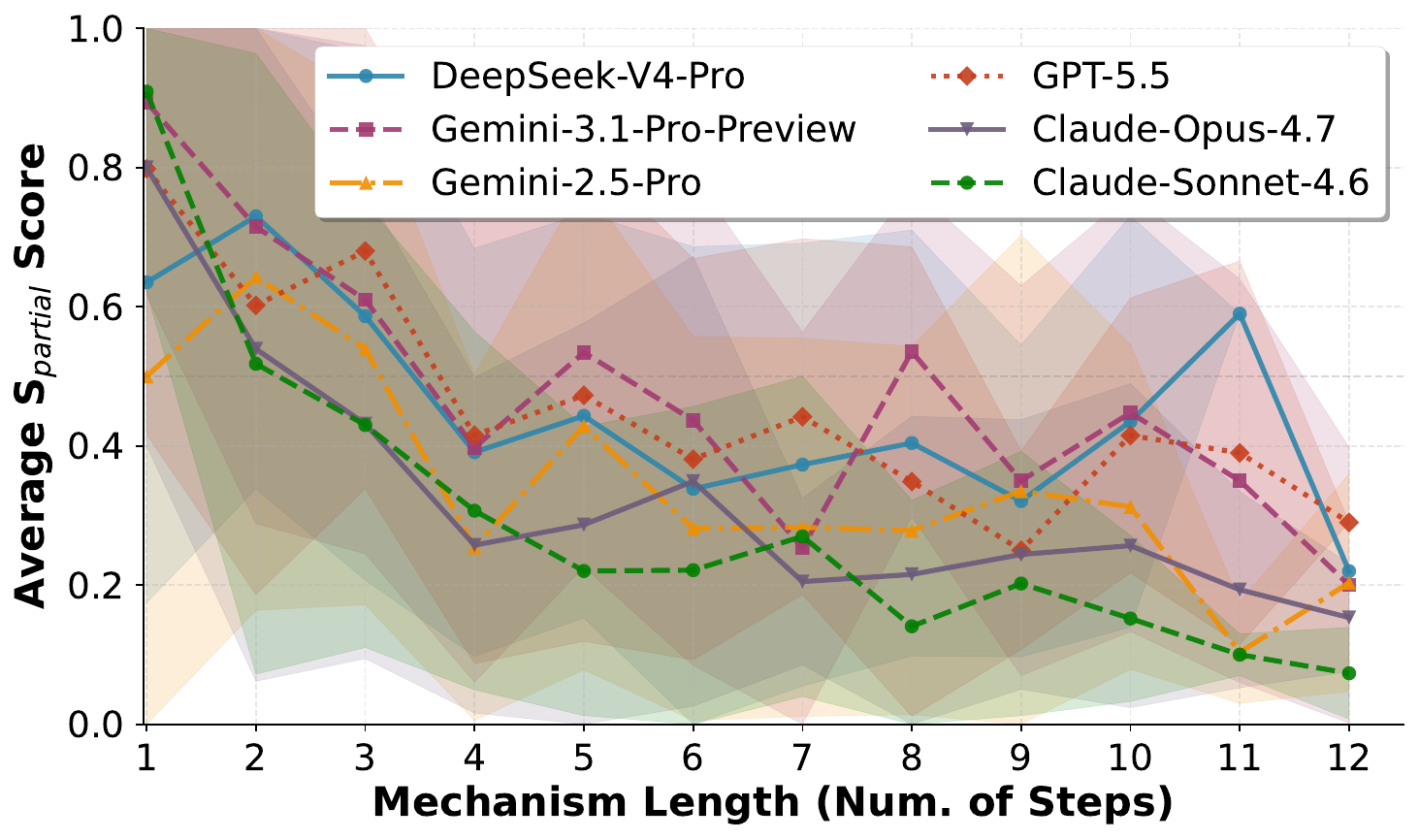}
    \vspace{-6mm}
    \caption{Performance vs Reaction Complexity. Performance degradation as length increases. Longer pathways amplify illogical jumps, showing models’ limited ability to sustain causal reasoning across extended steps.}
    \label{fig:score_vs_length}
    \vspace{-3mm}
\end{figure}

\noindent\textbf{Length effect of the mechanism.} 
Fig.~\ref{fig:score_vs_length} shows that performance consistently drops with increasing mechanism length. While short two- or three-step mechanisms can often be reproduced, mechanisms longer than five steps frequently omit intermediates or rearrangements,
leading to incomplete or illogical pathways. 
This highlights their inability to sustain multi-step causal reasoning over longer contexts. 
Skipping critical steps, accounting for 20.8\% of total errors, reflects the models' difficulty in sustaining multi-step causal reasoning.
An example of this behavior is in Tab.~\ref{tab:case2}.

\begin{table}[htb]
\centering

\small
\begin{tabularx}{\linewidth}{
  >{\arraybackslash\hsize=0.15\hsize}X
  >{\arraybackslash\hsize=1.55\hsize}X
  >{\arraybackslash\hsize=1.3\hsize}X
}
\toprule
\textbf{Step} & \textbf{Model Output} & \textbf{Comments} \\
\midrule
1 & \textbf{acid\_base\_proton\_transfer:} \texttt{CC(C1CC1)=N[N-]...} & Correct initial deprotonation. \\ \hdashline[0.8pt/2pt]
2 & \textbf{acid\_base\_proton\_transfer:} \texttt{[CH2-]C(C1CC1)=N[N-]...} & Redundant step: removes $\alpha$-H instead of cleaving. \\ \hdashline[0.8pt/2pt]
3 \newline ...& \textbf{leaving\_group\_elimination:}  \newline \texttt{C=C(C1CC1)N=N} &
     Fabricate electron flow and charge distribution. \\
\bottomrule
\end{tabularx}
\caption{Example where the model achieves the correct result via implausible steps. Full results in App.~\ref{app:case-study-4}}
\label{tab:case2}
\vspace{-5mm}
\end{table}

\begin{figure}[!htb]
    \centering
    \includegraphics[width=\linewidth]{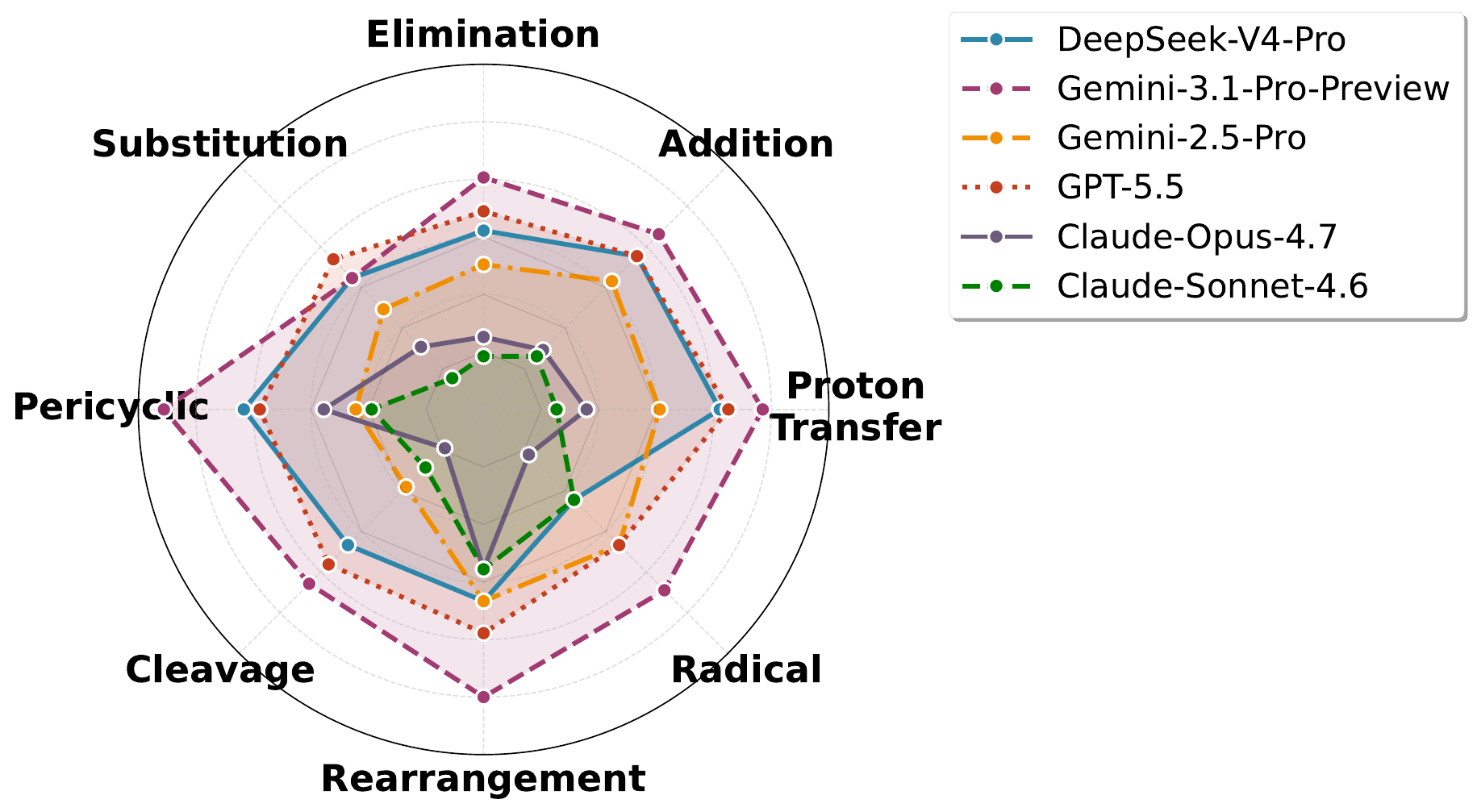}
    \caption{Model accuracy by type. While addition and substitution are handled moderately well, pericyclic, rearrangement, and radical steps remain challenging.}
    \label{fig:type_radar}
    \vspace{-5mm}
\end{figure}

\noindent\textbf{Inherently more challenging types.} 
Mechanism step types differ markedly in difficulty. Fig.~\ref{fig:type_radar} shows models perform better on frequent, pattern-based transformations such as substitution and addition, but struggle with rearrangements, pericyclic reactions, and radical processes. These categories demand deeper causal reasoning about electron flow, intermediate stability, and molecular structure, which are capabilities that current LLMs fail to develop. The observed disparity suggests that model predictions rely primarily on surface pattern association rather than mechanism understanding.
These results reveal two key deficiencies: a \textbf{reasoning gap}, where models fail to sustain coherent multi-step causal logic, and a \textbf{knowledge gap}, where limited mechanistic understanding leads to systematic errors in complex reaction types.



\label{sec:improvements}

\subsection{Improving LLM Performance}\label{sec:improving}
\noindent\textbf{In-context learning (ICL).}  
To narrow the \textbf{perceptual gap}, we apply a three-shot in-context learning strategy, selecting three reactions from the training corpus \textbf{oMe-Silver} with the highest \textit{Tanimoto similarity} of \textit{Differential Reaction Fingerprint} (DRFP)~\citep{probst2022reaction} to the query reaction. These exemplars provide structural priors that guide the model toward multi-step, causally coherent reasoning. As shown in Tab.~\ref{tab:icl_model_performance} and App.~\ref{app:case-study-icl}, exemplar-based prompting consistently improves scores across nearly all models. The gains are most pronounced for larger proprietary models, such as Claude-Sonnet-4.6, DeepSeek-V4-Flash. Improvements are especially strong for common step types, such as proton transfer and nucleophilic addition, where models can retrieve and adapt familiar mechanistic patterns. This suggests that ICL enhances reasoning not by introducing new knowledge, but by scaffolding the model’s retrieval and analogy capabilities. Further, the improvements also demonstrate the high quality of silver standard data.

To validate the necessity of DRFP-guided retrieval, we compare 
ICL-DRFP with randomly sampled demonstrations (ICL-Random) in 
Appendix~\ref{app:icl-ablation}. As shown in Table~\ref{tab:icl-ablation}, ICL-Random yields 
only marginal and often unstable gains. Random demonstrations 
help models follow the output format but frequently introduce 
irrelevant chemical contexts, leading to noisy or even negative 
effects. In contrast, ICL-DRFP produces substantial and 
consistent improvements across all models and metrics, 
confirming that mechanistically relevant exemplars act as 
effective structural priors for multi-step reasoning.


\noindent\textbf{Fine-tuning.}  
To move beyond exemplar imitation and further narrow the \textbf{reasoning} and \textbf{knowledge gaps}, we finetune open-source foundation models on our \textbf{oMe-Silver} dataset using supervised instruction tuning. All experiments are implemented in the \texttt{LLaMA-Factory} framework
~\citep{zheng2024llamafactory} and Transformers
~\citep{wolf-etal-2020-transformers} with BF16 precision on NVIDIA GH200 GPUs. 
Qwen-3-4B is trained with both standard SFT and Chain-of-Thought SFT (CoT-SFT). Their training data's only difference is that CoT-SFT additionally includes a natural-language rationale for each step. Details of the rationale construction and prompt are provided in App.~\ref{app:prompt_rationale}. LLaMA-3-8B, Mistral-7B, and OLMo-2-13B are optimized using parameter-efficient LoRA. Fine-tuning yields substantial improvements in all dimensions, particularly for in-domain reactions, with Qwen-3-SFT outperforming some proprietary models such as Claude-Sonnet-4.6 (Tab.~\ref{tab:fine-tune-optimized}).

Notably, incorporating CoT reasoning during SFT provides additional gains in mechanistic accuracy under ICL prompting. While ICL enhances reasoning by imitation, fine-tuning internalizes such reasoning patterns as stable model priors. Together, these results demonstrate that explicit mechanistic supervision is the most effective way to transfer mechanism knowledge and reasoning capabilities into open-source LLMs.

\section{Conclusion}

\label{sec:conclusion}
In conclusion, we present \textbf{oMeBench}, a benchmark for evaluating large language models on \emph{organic mechanism reasoning}. Evaluating on proprietary and open-source models, we identify persistent gaps in mechanistic understanding, particularly in multi-step causal reasoning and symbolic consistency. To address these challenges, we introduce \textbf{oMeS}, a dynamic evaluation framework, and show that both in-context learning and supervised fine-tuning enhance mechanistic reasoning, with our fine-tuned 4B model achieving comparable performance to Claude-Sonnet-4.6. Overall, oMeBench establishes a foundation for advancing chemically grounded reasoning in large language models.

\section*{Limitations}
We identify key limitations in our models, dataset, and evaluation framework. While several are discussed throughout the paper, we summarize the most critical aspects below for completeness

\paragraph{Models.}
Both our improved models and existing LLMs still struggle with mechanistic grounding. They sometimes misinterpret reaction conditions or generate chemically invalid intermediates, revealing limited integration between symbolic reasoning and molecular structure constraints. This indicates that current token-level generation lacks explicit enforcement of chemical conservation and contextual awareness.

\paragraph{Dataset.} While \textbf{oMeBench} includes graduate-level mechanisms across diverse reaction classes, its current scale remains smaller than large outcome-prediction corpora, limiting opportunities for large-scale pretraining.  Expanding the dataset to modern reaction classes, such as photoredox, enzymatic, and transition-metal catalysis, would enable broader and more realistic benchmarking. Moreover, while LLM-driven expansion introduces substantial structural diversity, it yields a skewed distribution of substituents; for example, CH3 is disproportionately overrepresented (Fig.~\ref{fig:group_cloud}), which may bias model learning toward common functional groups. 
Moreover, although we verified that Silver reactions do not overlap with Gold (App.~\ref{app:contamination}), LLM-based expansion may still inherit subtle biases from the generator model. Exploring non-LLM-based methods for silver-standard data construction is an important direction for future work.

\paragraph{Evaluation.} We recruit annotators with strong chemistry expertise (see Appendix~\ref{app:annotator}), whose judgments may differ from those of general or non-expert annotators.
Although \textbf{oMeS} offers a fine-grained and chemically grounded evaluation in both molecular and mechanistic level, it still abstracts molecules into symbolic SMILES representations, which cannot fully capture three-dimensional environments. Incorporating graph- or geometry-based similarity measures could yield a more complete assessment of reasoning accuracy.


\section*{Ethical Considerations}

\paragraph{Data.} All data were obtained from publicly available textbooks, peer-reviewed publications, and open-access reaction databases, in full compliance with relevant legal and ethical standards. No proprietary or confidential sources were used. 

\paragraph{Annotation.} Human annotations were carried out by qualified chemistry researchers, compensated at the standard institutional rate; anonymized summaries of their expertise and experience are provided in Appendix~\ref{app:annotator}.

\paragraph{Model and Environmental Ethics.}
All models used are open-source or publicly available under research-friendly licenses. Because our methods are based on pretrained language models, they may inherit biases from their underlying training data. Moreover, since most chemical sources are written in English, certain linguistic or regional chemistries may be underrepresented. We also recognize the environmental costs associated with large-scale model computation and therefore restrict our work to evaluation and lightweight fine-tuning.

\paragraph{Responsible Research.}
We aim to promote fairness, transparency, and reproducibility in AI-for-science research. The tools and benchmarks introduced in this work are designed to assist, not replace, human chemists. All results should be interpreted with expert judgment.


 


\subsection*{Use Of Generative AI.}
We employed large language models (e.g., ChatGPT) to assist with grammar refinement, LaTeX code formatting, and, in a few cases, improving the conciseness of lengthy sentences in the authors’ original handwritten content.
All outputs were subsequently reviewed and critically revised by the human authors to ensure scientific accuracy and originality.

\newpage

\bibliography{rebib}

\appendix

\newpage
\section{Annotation Detail}
\label{app:annotator}
\paragraph{Annotator Background.}  
We recruited four expert annotators with strong backgrounds in chemistry competitions. Two are gold medalists of the Chemistry Olympiad, while the other two are regional first-prize winners. All annotators have received more than four years of rigorous training in organic chemistry, including functional group transformations, reaction mechanisms, stereochemistry, and common strategies in synthetic design, and are currently pursuing degrees in chemistry. To ensure both competence and reliability, the selection criteria required (i) at least a first prize in chemistry competitions and enrollment in a relevant major, (ii) demonstration of responsibility and enthusiasm during interviews, and (iii) successful completion of pilot annotations that reflected correct understanding of the task.

\paragraph{Annotation Procedure.}  
The annotation process began with the authors curating candidate reactions and defining preliminary criteria for reaction \textit{type}, \textit{subtype}, and \textit{difficulty level}. Annotators were first introduced to the project background and refreshed their knowledge of key organic chemistry concepts before receiving detailed task instructions. They were encouraged to refine and clarify the labeling criteria through iterative discussions with the authors. The core tasks involved (i) extracting reactions and validating intermediate SMILES strings, and (ii) annotating metadata such as difficulty level, type, and subtype. In later stages, annotators also contributed to labeling error categories and drafting representative case studies.

\paragraph{Inter-Annotator Agreement.}  
Each reaction was independently annotated by two domain experts, and disagreements were resolved through consultation with a third annotator followed by consensus discussion. To quantify annotation consistency, we measured inter-annotator agreement across major labeling dimensions. Fleiss’ $\kappa$ for reaction types reached 0.95, indicating almost perfect agreement, while the quadratic-weighted $\kappa$ for difficulty levels was 0.86, reflecting strong consistency across annotators. These results demonstrate that the annotation protocol yields highly reliable and reproducible human labels, providing a stable foundation for subsequent similarity-based evaluation.

\subsection{Instructions to Expert Annotators}

\paragraph{Task Definition.}
Given a \textbf{reaction entry} with its \texttt{reactants\_smiles}, \texttt{products\_smiles}, and \texttt{conditions}, your task is to construct or verify a complete and correct \textbf{mechanism}. Each entry must include all required metadata fields, ensuring machine readability and chemical validity. The target JSON schema is as follows:

\begin{verbatim}
{
 "reaction_id": "XX-xxx",
 "level": "easy/medium/hard",
 "name": "...",
 "reactants_smiles": [...],
 "products_smiles": [...],
 "conditions": "...",
 "mechanism_step_nums": 4,
 "description": "...",
 "mechanism": [...]
}
\end{verbatim}

\paragraph{Required Metadata Fields.}
Annotators must carefully fill out or verify the following components:

\begin{enumerate}
    \item \textbf{reaction\_id:} A unique alphanumeric identifier (e.g., “NR-201”).  
    \item \textbf{level:} Difficulty level: \textit{easy}, \textit{medium}, or \textit{hard}, based on mechanistic complexity and reasoning depth.  
    \item \textbf{name:} The formal or commonly accepted name of the reaction (e.g., “Nazarov Cyclization Reaction”).  
    \item \textbf{reactants\_smiles / products\_smiles:} Chemically valid SMILES strings representing all reactants and products.  
    \item \textbf{conditions:} Concise expression of reaction conditions, including catalysts, reagents, or solvents (e.g., “H\textsuperscript{+} OSO\textsubscript{2}Me”).  
    \item \textbf{description:} A short natural-language summary of the overall mechanism and rationale.  
    \item \textbf{mechanism:} A list of ordered elementary steps. Each step must contain:  
    \begin{compactitem}
        \item \texttt{step}: Step number (integer, starting from 1).  
        \item \texttt{type}: The high-level mechanistic category (e.g., \textit{addition}, \textit{elimination}, \textit{proton\_transfer}).  
        \item \texttt{subtype}: The specific mechanistic subtype (e.g., \textit{acid\_base\_proton\_transfer}, \textit{nucleophilic\_addition}).  
        \item \texttt{intermediate\_smiles}: A chemically valid SMILES representing the intermediate after this step.  
        \item \texttt{rationale}: A brief expert-level explanation of the mechanistic principle underlying this transformation (e.g., “Protonation of the carbonyl oxygen increases electrophilicity and stabilizes the developing positive charge via resonance.”)  
    \end{compactitem}
\end{enumerate}

\paragraph{Annotation Principles.}
\begin{enumerate}[leftmargin=1.5em]
    \item Each mechanism must align with those recordings in literature, textbooks or databases.
    \item All SMILES strings, including reactants, products and intermediates, must satisfy chemical validity and are c\textbf{ompilable by RDKit}.
    \item Assign step types and subtypes according to the standardized taxonomy provided in Tab.~\ref{tab:types_subtypes_grouped}.
    \item Annotators should provide rationales and descriptions in \textbf{concise academic language}, emphasizing reactivity, resonance, or stability reasoning rather than descriptive repetition.
\end{enumerate}

\onecolumn
\paragraph{Example Annotation.}
\begin{jsonbox}
{
  "reaction_id": "NR-201",
  "level": "medium",
  "name": "Nazarov Cyclization Reaction",
  "reactants_smiles": ["C(C)=CC(=O)C=C(C)", "CS(=O)(=O)O"],
  "products_smiles": ["CC1=CC(=O)CC1(C)"],
  "conditions": "H+ OSO2Me",
  "mechanism_step_nums": 4,
  "description": "Under strong acid conditions, the divinyl ketone is protonated to activate the carbonyl toward cation formation. This enables oxyallyl cation generation, which undergoes a Nazarov 4$\pi$-electrocyclization to form a cyclopentenyl cation. Subsequent deprotonation and tautomerization yield the thermodynamically stable cyclopentenone product.",
  "mechanism": [
    {
      "step": 1,
      "type": "proton_transfer",
      "subtype": "acid_base_proton_transfer",
      "intermediate_smiles": "C(C)=CC(=[OH+])C=CC",
      "rationale": "Protonation of the carbonyl oxygen increases electrophilicity and stabilizes the developing positive charge via resonance, facilitating oxyallyl cation formation under strong acid."
    },
    ...
}
\end{jsonbox}
\twocolumn

\section{Dataset}
\subsection{Related Work}
\label{app:related_work}
A wide range of datasets have been developed to support machine learning in chemistry. General-purpose chemical databases such as Pistachio \citep{mayfield2018pistachio} and PubChem \citep{kim2025pubchem} provide millions of reactions and molecular entries compiled from ELNs, patents, and scientific publications. While valuable for large-scale data mining, these resources mainly emphasize reaction outcomes or compound properties, without explicit mechanistic details.

Among reaction-focused datasets, the USPTO collections are the most widely used. USPTO-50k \citep{lowe2012extraction} and USPTO-MIT \citep{jin2017predicting} provide thousands of reactions extracted from patents, but they primarily encode transformations from reactants to products without mechanistic rationales. Later variants such as mech-USPTO-31k \citep{chen2024large} and USPTO-Mechanistic \citep{joung2024reproducing} introduced template-based mechanism generation. However, these datasets are generated by automated matching procedures with atom-mapped SMILES, resulting in limited mechanistic diversity and repetitive patterns. They fail to reflect the stepwise reasoning chemists employ when analyzing electron flow and intermediates. 

Other efforts have attempted to collect mechanistic steps more explicitly. RMechDB \citep{tavakoli2023rmechdb} focuses on radical reactions, aggregating over 5,000 manually curated steps linked to specific transition states. Its counterpart, PMechDB, specializes in polar reactions \citep{tavakoli2024pmechdb}, providing a complementary collection of elementary steps. While valuable, both databases focus narrowly on single mechanistic families, and do not integrate radical and ionic logic in unified pathways. Along with USPTO family, their step collections remain fragmented, making them less suitable for benchmarking end-to-end reasoning. Moreover, this mapping representation is suitable for training neural networks, e.g. GNN, but far from a good dataset for LLM fine-tuning and evaluation.

Recently, datasets like ChemCoTBench \citep{li2025beyond} and SciBench \citep{wang2024scibenchevaluatingcollegelevelscientific} introduced natural language chemistry tasks, including multiple-choice questions or descriptive reasoning. SciBench only contains general chemical QA. As for ChemCoTBench, besides its low difficulty, it primarily frames chemical mechanisms as a few-step, simple multiple-choice questions, without modeling the multi-step causal structure of organic mechanisms. These benchmarks emphasize natural language interaction but lack explicit molecule-level mechanism details, limiting their ability to test fine-grained chemical reasoning. A detailed comparison is shown in Table~\ref{tab:benchmark_comparison}. 

In summary, existing datasets fall into two extremes: large-scale outcome-oriented corpora without mechanisms, or narrowly focused collections that lack breadth and natural language interpretability. Our benchmark complements these efforts by combining stepwise mechanistic detail, validated intermediates, and natural language rationales in a unified, diverse, and challenging dataset.

\onecolumn
\begin{table*}[htbp]
\centering
\small
\begin{tabularx}{\linewidth}{l c c c X}
\toprule
\textbf{Dataset} & \textbf{Form} & \textbf{\# Mechanism Items} & \textbf{\# Reaction Types} & \textbf{Sample} \\
\midrule

SciBench & QA & 30 & 0 & 
Calculate the energy of a photon for a wavelength of 100 pm about one atomic diameter. \\
\\
ChemCoTBench & MCQ & 100 & 50 & 
For reaction class: `CO$_2$H-Me deprotection', under the condition of `Deprotection with TMSI' and given reagents (written in SMARTS format) `[Si]-[I]', which following description is the correct elementary reaction stages description, considering the mechanism of this type of reaction? Choices: A: Ester and TMSI reaction $\mid$ Iodide attack $\mid$ Quench; B: Hydroxide addition $\mid$ Alkoxide leaves; C: Water addition $\mid$ Deprotonation $\mid$ Alkoxide leaves. \\
\\
Ours & Reasoning & 2000+ & 196 & 
See paper. \\

\bottomrule
\end{tabularx}
\caption{Comparison with existing chemistry reasoning benchmarks.}
\label{tab:benchmark_comparison}
\end{table*}

\subsection{Data format and comparison}
\label{app:data_format_comparion}
\subsubsection{mech-USPTO-31k Sample}
Below we show the data format in mech-USPTO-31k~\cite{chen2024large}, corresponding to a single line in the CSV file. We can see that early datasets primarily rely on \textbf{atom-mapping} to represent individual mechanistic steps, which is non-intuitive and difficult to interpret. Moreover, in the raw CSV format, it is difficult to determine which mechanistic steps together constitute a complete reaction.

\begin{tcolorbox}[title=A mechanistic step in mech-USPTO-31k]
Original Reaction: 
\begin{lstlisting}
O=C(OCc1ccccc1)[NH:1][CH2:2][CH2:3][CH2:4][CH2:5][C@@H:6]([C:7]([O:8][CH3:9])=[O:10])[NH:11][C:12](=[O:13])[NH:14][c:15]1[cH:16][c:17]([O:18][CH3:19])[cH:20][c:21]([C:22]([CH3:23])([CH3:24])[CH3:25])[c:26]1[OH:27]>>[NH2:1][CH2:2][CH2:3][CH2:4][CH2:5][C@@H:6]([C:7]([O:8][CH3:9])=[O:10])[NH:11][C:12](=[O:13])[NH:14][c:15]1[cH:16][c:17]([O:18][CH3:19])[cH:20][c:21]([C:22]([CH3:23])([CH3:24])[CH3:25])[c:26]1[OH:27] 
\end{lstlisting}
Updated Reaction: 
\begin{lstlisting}
[O:201]=[C:101]([O:202][CH2:203][c:204]1[cH:205][cH:206][cH:207][cH:208][cH:209]1)[NH:1][CH2:2][CH2:3][CH2:4][CH2:5][C@@H:6]([C:7]([O:8][CH3:9])=[O:10])[NH:11][C:12](=[O:13])[NH:14][c:15]1[cH:16][c:17]([O:18][CH3:19])[cH:20][c:21]([C:22]([CH3:23])([CH3:24])[CH3:25])[c:26]1[OH:27].[BrH:301]>>[NH2:1][CH2:2][CH2:3][CH2:4][CH2:5][C@@H:6]([C:7]([O:8][CH3:9])=[O:10])[NH:11][C:12](=[O:13])[NH:14][c:15]1[cH:16][c:17]([O:18][CH3:19])[cH:20][c:21]([C:22]([CH3:23])([CH3:24])[CH3:25])[c:26]1[OH:27] 
\end{lstlisting}
Mechanistic Class: \textit{Cbz\_deprotection}

Mechanistic Label: 
\begin{lstlisting}
[(201, 301.1), ([301.1, 301], 301), (301, 203), ([203, 202], [202, 101]), ([101, 201], 201), (1, 201.1), ([201.1, 201], 201), (201, [201, 101]), ([101, 1], 1)]
\end{lstlisting}
\end{tcolorbox}

\subsubsection{oMe-Gold Sample}
Below, we present the data format used in our datasets. It provides clear reaction information, including SMILES of reactants, products, and conditions, as well as a step-by-step mechanism with annotations of type, subtype, and intermediate SMILES. In addition, each step is accompanied by fine-grained rationales, and each reaction is accompanied by a natural language description explaining the underlying reasons for the transformations. The dataset is stored in JSON/JSONL format, enabling both large-scale evaluation and fine-tuning, while remaining interpretable for humans and domain experts.

\begin{jsonbox}
{
  "reaction_id": "NR-201",
  "level": "medium",
  "name": "Nazarov Cyclization Reaction",
  "reactants_smiles": ["C(C)=CC(=O)C=C(C)", "CS(=O)(=O)O"],
  "products_smiles": ["CC1=CC(=O)CC1(C)"],
  "conditions": "H+ OSO2Me",
  "mechanism_step_nums": 4,
  "description": "Under strong acid conditions, the divinyl ketone is protonated to activate the carbonyl toward cation formation. This enables oxyallyl cation generation, which undergoes a Nazarov 4$\pi$-electrocyclization to form a cyclopentenyl cation. Subsequent deprotonation and tautomerization yield the thermodynamically stable cyclopentenone product.",
  "mechanism": [
    {
      "step": 1,
      "type": "proton_transfer",
      "subtype": "acid_base_proton_transfer",
      "intermediate_smiles": "C(C)=CC(=[OH+])C=CC",
      "step_weight": 0.1020,
      "rationale": "Protonation of the carbonyl oxygen increases electrophilicity and stabilizes the developing positive charge via resonance, facilitating oxyallyl cation formation under strong acid."
    },
    {
      "step": 2,
      "type": "pericyclic",
      "subtype": "electrocyclization",
      "intermediate_smiles": "C1(=CC(C([CH+]1)C)C)O",
      "step_weight": 0.5714,
      "rationale": "The oxyallyl cation undergoes a conrotatory 4$\pi$-electrocyclization, favored by orbital symmetry and leading to a cyclopentenyl cation stabilized by adjacent substituents."
    },
    {
      "step": 3,
      "type": "elimination",
      "subtype": "proton_elimination",
      "intermediate_smiles": "C1(=CC(C(=C1)C)C)O",
      "step_weight": 0.2449,
      "rationale": "Deprotonation relieves positive charge on the cyclopentenyl cation, driven by formation of a more conjugated alkene system."
    },
    {
      "step": 4,
      "type": "proton_transfer",
      "subtype": "acid_base_proton_transfer",
      "intermediate_smiles": "CC1=CC(=O)CC1(C)",
      "step_weight": 0.0816,
      "rationale": "Acid-catalyzed tautomerization converts the enol to the more stable conjugated ketone, favored by carbonyl bond strength and extended $\pi$-conjugation."
    }
  ]
}
\end{jsonbox}

\subsubsection{oMe-Template Sample}
Substitutable R-groups are denoted as \texttt{[*:i]} for RDKit compatibility. 

\begin{jsonbox}
{
  "reaction_id": "NR-201",
  "level": "medium",
  "name": "Nazarov Cyclization Reaction",
  "reactants_smiles": ["C(C([*:1]))=CC(=O)C=C(C([*:2]))", "CS(=O)(=O)O"],
  "products_smiles": ["C([*:2])C1=CC(=O)CC1([*:1])"],
  "conditions": "H+ OSO2Me",
  "mechanism_step_nums": 4,
  "description": "The reaction begins with protonation of the carbonyl oxygen in the divinyl ketone, forming a resonance-stabilized oxyallyl cation. This cation undergoes a 4$\pi$-electrocyclic ring closure to yield a cyclopentenyl cation intermediate, followed by deprotonation and keto-enol tautomerization to produce the cyclopentenone product.",
  "mechanism": [
    {
      "step": 1,
      "type": "proton_transfer",
      "subtype": "acid_base_proton_transfer",
      "intermediate_smiles": "C(C([*:1]))=CC(=[OH+])C=CC([*:2])",
      "step_weight": 0.1020,
      "rationale": "Protonation of the carbonyl oxygen activates the divinyl ketone toward cation formation by stabilizing the developing positive charge."
    },
    {
      "step": 2,
      "type": "pericyclic",
      "subtype": "electrocyclization",
      "intermediate_smiles": "C1(=CC(C([CH+]1)C([*:1]))C([*:2]))O",
      "step_weight": 0.5714,
      "rationale": "The oxyallyl cation undergoes a conrotatory 4$\pi$-electrocyclization, generating a cyclopentenyl cation stabilized by conjugation and substituent effects."
    },
    {
      "step": 3,
      "type": "elimination",
      "subtype": "proton_elimination",
      "intermediate_smiles": "C1(=CC(C(=C1)C([*:2]))C([*:1]))O",
      "step_weight": 0.2449,
      "rationale": "Deprotonation of the cation intermediate restores aromatic conjugation and reduces charge localization."
    },
    {
      "step": 4,
      "type": "proton_transfer",
      "subtype": "acid_base_proton_transfer",
      "intermediate_smiles": "C([*:2])C1=CC(=O)CC1(C([*:1]))",
      "step_weight": 0.0816,
      "rationale": "Final tautomerization converts the enol to the stable conjugated ketone, completing the Nazarov cyclization."
    }
  ]
}
\end{jsonbox}

\subsection{Types and Subtypes}
\begin{table}[H]
\centering
\small

\begin{tabular}{@{}l l p{9.2cm}@{}}
\toprule
\textbf{Type} & \textbf{Subtype} & \textbf{Description} \\
\midrule

\multirow{2}{*}{cleavage}
  & heterolytic\_cleavage & Bond breaks with both electrons going to one fragment (ions formed), including decarboxylation in some condition \\
  & homolytic\_cleavage & Bond breaks evenly with one electron to each fragment (radicals formed) \\

\midrule
\multirow{3}{*}{addition}
  & nucleophilic\_addition & Nucleophile attacks an electrophilic center \\
  & electrophilic\_addition & Electrophile adds to an electron-rich system (e.g., alkene) \\
  & radical\_addition & Radical adds to a $\pi$-bond system \\

\midrule
\multirow{3}{*}{elimination}
  & proton\_elimination & $\beta$-H proton of carbocation is abstracted to form a double bond \\
  & leaving\_group\_elimination & Leaving group departs simultaneously with double bond formation \\
  & radical\_elimination & Radical eliminates a small fragment to form a $\pi$-bond \\

\midrule
\multirow{3}{*}{substitution}
  & nucleophilic\_substitution & Nucleophile replaces a leaving group \\
  & electrophilic\_substitution & Electrophile replaces a hydrogen atom on an aromatic ring \\
  & radical\_substitution & Radical replaces a group within a $\sigma$-bond \\

\midrule
\multirow{2}{*}{rearrangement}
  & 1,2-shift & 1,2-alkyl or hydride shift \\
  & radical\_rearrangement & Radical migrates to a different position \\

\midrule
proton\_transfer & acid\_base\_proton\_transfer & Proton migrates between a donor and an acceptor \\

\midrule
electron\_transfer & single\_electron\_transfer & Single electron moves from donor to acceptor \\

\midrule
coordination & lewis\_acid\_base\_coordination & Lone pair coordinates with an empty orbital \\

\midrule
\multirow{4}{*}{radical}
  & radical\_initiation & Generation of radicals (e.g., by light, heat) \\
  & radical\_propagation & Radical continuously generates new radicals \\
  & radical\_termination & Two radicals combine to terminate chain reactions \\
  & radical\_coupling & Two radicals form a new bond directly (e.g., C--C or C--N) \\

\midrule
\multirow{6}{*}{pericyclic}
  & cycloaddition & Two $\pi$-systems combine to form a ring (e.g., Diels--Alder) \\
  & electrocyclization & Ring closure or opening via conjugated $\pi$-system rotation \\
  & sigmatropic\_rearrangement & Migration of $\sigma$-bond across a $\pi$-system (e.g., [3,3]-Claisen) \\
  & group\_transfer & Group is transferred via concerted cyclic pathway \\
  & ene\_reaction & Transfer of an allylic hydrogen and formation of a new $\sigma$-bond \\
  & cheletropic\_reaction & Two new bonds formed to the same atom (subset of cycloaddition) \\

\bottomrule
\end{tabular}
\caption{Reaction Types, Subtypes, and Descriptions}
\label{tab:types_subtypes_grouped}

\end{table}

\newpage
\subsection{Data Contamination Analysis}
\label{app:contamination}
We take two measures to prevent contamination between the Silver training set and the Gold evaluation set.

\noindent\textbf{Exact-match filtering.} During oMe-Silver construction (App.~\ref{app:scaling_algorithm}), we remove any generated reaction whose canonical SMILES exactly matches a reaction in oMe-Gold.

\noindent\textbf{Nearest-neighbor similarity scan.} To rule out near-duplicates, we compute reaction-level Morgan fingerprints (radius = 2, 2048 bits) for all Silver--Gold pairs and measure their Tanimoto similarity. Our analysis shows that no pair exceeds a similarity of 0.80.

\begin{figure*}[hb]
    \centering
    \includegraphics[width=0.95\linewidth]{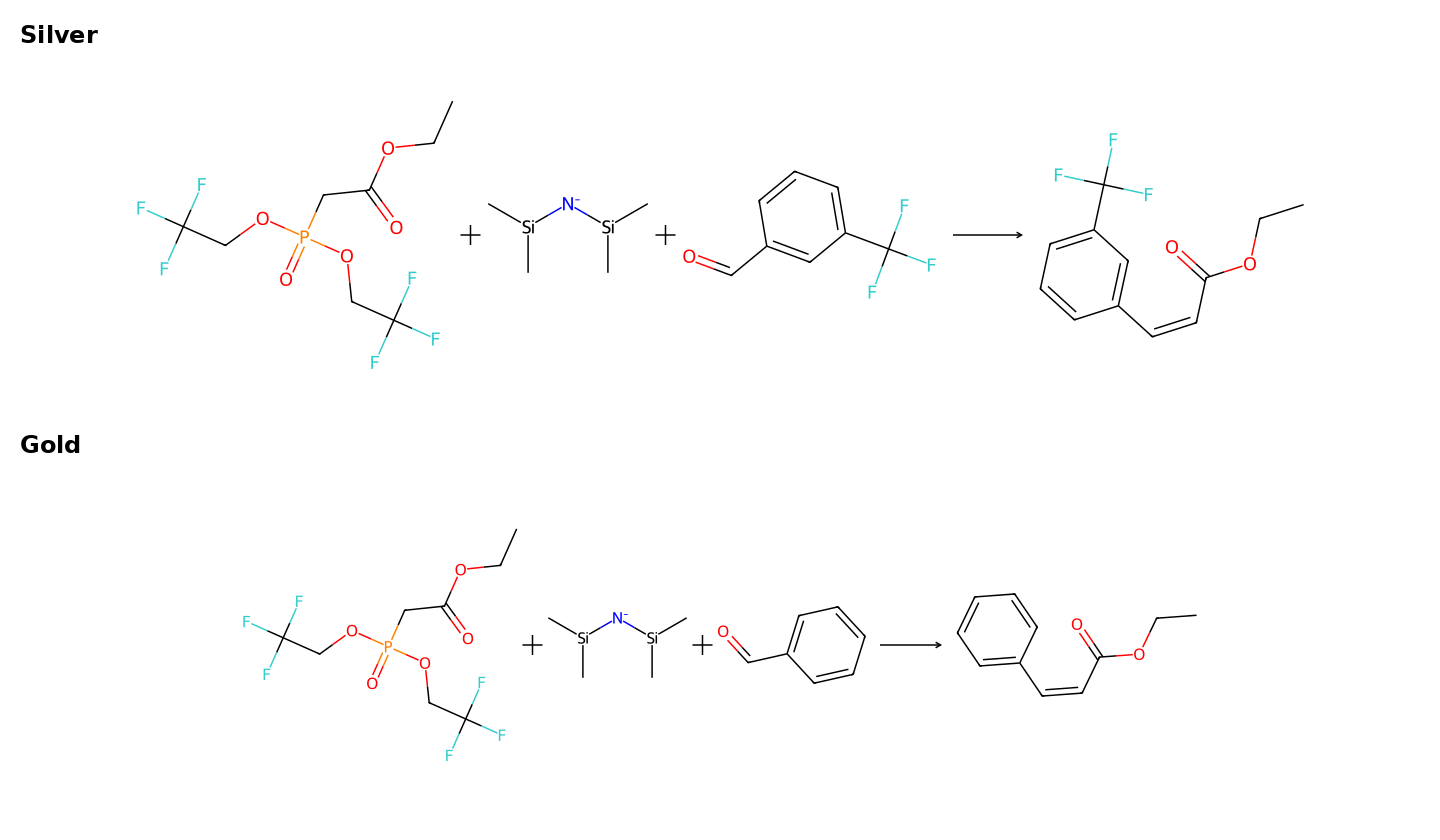}
    \caption{Highest-similarity Silver--Gold pair (Tanimoto = 0.7969). The two reactions share the same phosphonate reagent, HMDS base, and Horner--Wadsworth--Emmons transformation, but differ in the aldehyde substrate (3-CF$_3$-benzaldehyde vs.\ benzaldehyde), yielding different products.}
    \label{fig:contamination_example}
\end{figure*}

The highest observed similarity (0.7969) occurs between the following pair, where the aldehyde substrate differs by a substantial CF$_3$ substituent (Fig.~\ref{fig:contamination_example}).

Note that moderate fingerprint similarity is expected between chemically valid reactions of the same mechanistic class: in organic mechanisms, structural changes between two valid reactions cannot be arbitrarily large without altering the mechanism itself. True contamination would require near-identical reactions, which we do not observe. This confirms that Silver reactions do not duplicate or closely shadow any Gold reactions, indicating negligible risk of contamination.

\newpage
\onecolumn
\section{Weighting}
\subsection{Weighting Strategy}
\label{app:weighting_strategy}

Concretely, each step's raw weight combines: (i) a base weight by step \emph{type} (e.g., pericyclic $>$ rearrangement $>$ elementary proton transfer), (ii) a \emph{subtype} modifier (e.g., sigmatropic or cheletropic steps are up-weighted relative to baseline nucleophilic additions), and (iii) contextual adjustments:
\begin{itemize}
    \item \textbf{Positional adjustment} $P_{\text{position}}$: terminal steps that mainly "copy" product structure are slightly down-weighted;
    \item \textbf{Structural bonus} $B_{\text{special}}$: ring closures, key bond-forming events, etc., receive small positive bonuses.
\end{itemize}
The raw weight for step $i$ is
\[
w_i^{\text{raw}} \;=\; \text{clip}\Big( W_{\text{type}} \times M_{\text{subtype}} \times P_{\text{position}} \;+\; B_{\text{special}},\ 0.5,\ 6.0 \Big),
\]
and the final weights are normalized: $w_i = w_i^{\text{raw}} \big/ \sum_{k=1}^{N} w_k^{\text{raw}}$. This scheme (curated by three domain experts) mirrors how chemists prioritize steps and ensures stability while preserving contrast between trivial and non-trivial reasoning.

\subsection{Weighting Implementation}
\label{sec:weighting-code}

For reproducibility, we summarize in Table~\ref{tab:weighting-rules} the weighting configuration used in our benchmark. The final step weight is computed by combining these factors as described in Section~4.2 (Metrics).

\begin{table}[htbp]
\centering
\resizebox{\textwidth}{!}{
\begin{tabular}{@{}llll@{}}
\toprule
\textbf{Factor} & \textbf{Category} & \textbf{Values} & \textbf{Notes} \\
\midrule
\multirow{10}{*}{Base type weights} 
& Pericyclic & 4.0 & Highest complexity (cycloadditions, sigmatropic shifts) \\
& Rearrangement & 3.5 & Includes 1,2- and 1,3-shifts \\
& Radical & 3.0 & Radical initiation, propagation, coupling \\
& Electron transfer & 3.0 & Single- or multi-electron processes \\
& Substitution & 2.5 & Nucleophilic/electrophilic substitutions \\
& Cleavage & 2.5 & Heterolytic/homolytic bond cleavage \\
& Addition & 2.0 & Nucleophilic/electrophilic additions \\
& Elimination & 2.0 & Deprotonation, leaving-group elimination \\
& Coordination & 1.5 & Metal-ligand binding, weak interactions \\
& Proton transfer / Dissociation & 1.0 & Elementary steps, down-weighted \\
\midrule
\multirow{8}{*}{Subtype modifiers} 
& Sigmatropic rearrangement & 1.3 & Extra difficulty \\
& Cheletropic reaction & 1.3 & Extra difficulty \\
& Cycloaddition / Ene & 1.2 & Ring closure / pericyclic steps \\
& Homolytic cleavage & 1.2 & More complex than heterolytic \\
& Electrophilic substitution & 1.2 & Context-dependent selectivity \\
& Nucleophilic substitution & 1.1 & \\
& Leaving-group elimination & 1.1 & \\
& Proton transfer (acid-base) & 0.9 & Simplest step, discounted \\
\midrule
Positional modifier & Last step & 0.9 & Often bookkeeping proton transfers \\
\midrule
\multirow{3}{*}{Special bonuses $B_{\text{special}}$} 
& Ring closure (``cycl" in subtype) & +0.3 & \\
& Key bond-forming step (addition/substitution) & +0.2 & Central to overall mechanism \\
\bottomrule
\end{tabular}
}
\caption{Weighting configuration for mechanistic steps.}
\label{tab:weighting-rules}
\end{table}

\newpage

\section{Scaling Algorithm}
\label{app:scaling_algorithm}

We scale \textbf{oMe-Template} into \textbf{oMe-Silver} using a generative augmentation procedure as the pseudocode outlined below.  

\begin{algorithm}[h]
\caption{Mechanism Template Up-Scaling}
\label{alg:scaling}
\begin{algorithmic}[1]
\Require Template set \textbf{oMe-Template} $\mathcal{T} = \{T_1, \dots, T_n\}$ , LLM $\mathcal{G}$ , Validator $\mathcal{V}$ (e.g., RDKit)
\Ensure Valid mechanism dataset \textbf{oMe-Silver} $\mathcal{D}_{\text{aug}}$

\State Initialize $\mathcal{D}_{\text{aug}} \gets \emptyset$ 
\For{each template $T \in \mathcal{T}$}
    \State $R \gets$ \Call{ExtractRGroups}{$T$} \Comment{Identify reactive sites / substitutable groups}
    \State $S \gets$ \Call{QueryLLM}{$\mathcal{G}, T, R$} \Comment{Use LLM to propose candidate R-groups}
    \State $C \gets$ \Call{EnumerateCombinations}{$R, S$} \Comment{Generate all feasible substitution combinations}
    \For{each combination $c \in C$} 
        \State $T_c \gets$ \Call{SubstituteRGroups}{$T, c$} \Comment{Construct a new mechanism instance}
        \If{\Call{IsValid}{$\mathcal{V}, T_c$}}
            \State $\mathcal{D}_{\text{aug}} \gets \mathcal{D}_{\text{aug}} \cup \{T_c\}$ \Comment{Add valid mechanism to dataset}
        \EndIf
    \EndFor
\EndFor
\State \Return $\mathcal{D}_{\text{aug}}$
\end{algorithmic}
\end{algorithm}

\begin{figure}[!htb]
    \centering
    \includegraphics[width=0.5\linewidth]{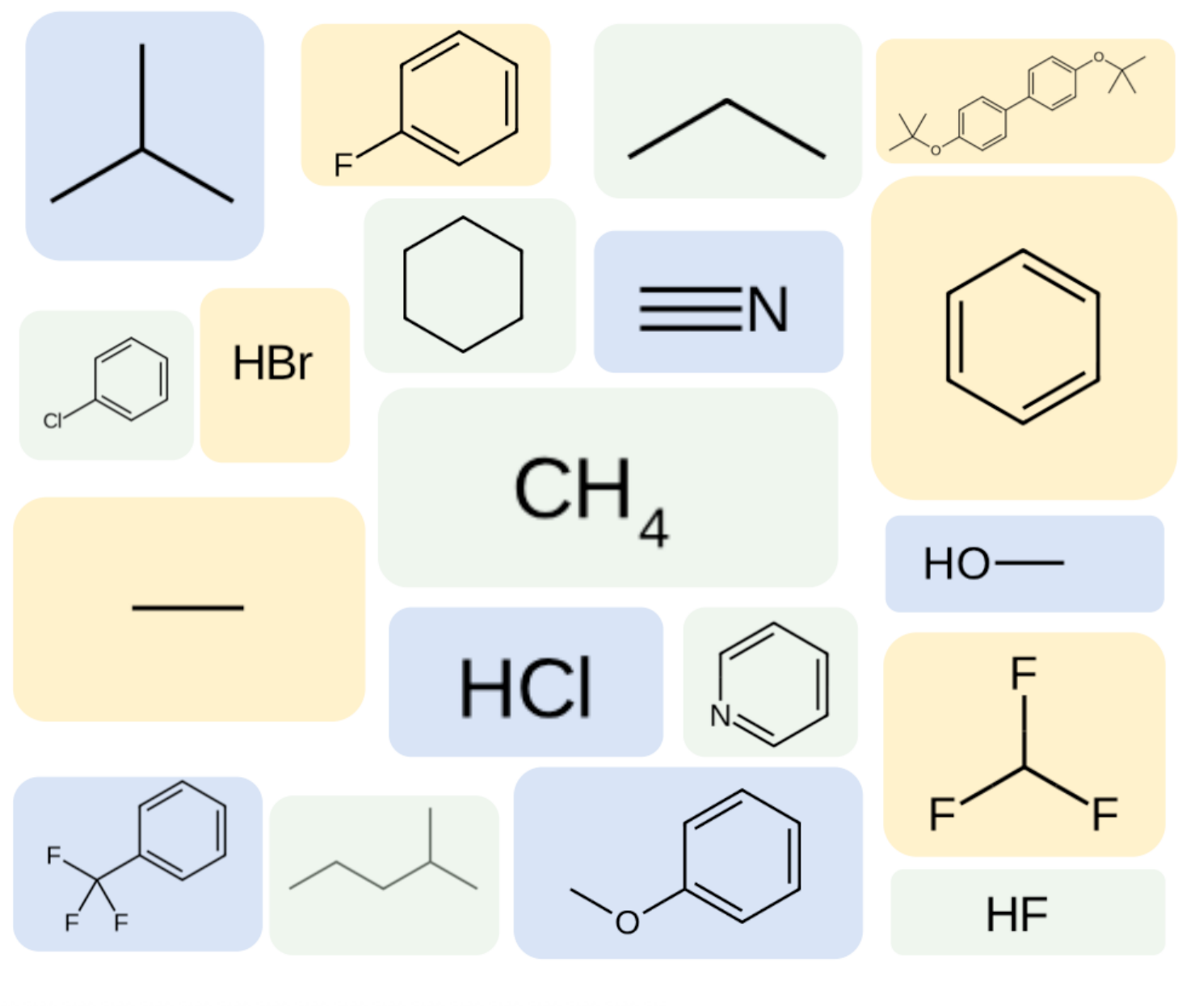}
    \caption{Top 19 R-groups most frequently suggested by Gemini-Pro-2.5}
    \label{fig:group_cloud}
\end{figure}

\section{Metrics Details}
\label{app:metrics_details}
The formula for $V$ and $L$ are defined as follows: 
\[
V \;=\; \frac{1}{M}\sum_{j=1}^{M} \mathbf{1}\big[\text{RDKit parses } \hat I_j\big].
\]

\[
L \;=\; \frac{1}{N}\sum_{k=1}^{N} \mathbf{1}\big[t_k = \hat t_k\big],
\]
where a missing alignment at position $k$ counts as incorrect.

\subsection{Robustness to Similarity Choice}
\label{app:similarity_ablation}

To verify that \textbf{oMeS} does not depend on a specific similarity function, we re-compute $S_{part}$ using GESim~\citep{shiokawa2025gesim}, a graph-based similarity measure built on von Neumann graph entropy. GESim captures global molecular topology (ring scaffold, connectivity), in contrast to Tanimoto on Morgan fingerprints, which focuses on local functional-group and bond-level differences.

Table~\ref{tab:similarity_ablation} reports $S_{part}$ under both metrics. Although GESim yields higher absolute scores due to its focus on global topology, the relative ranking of all 13 models is preserved, with a Spearman correlation of $\rho = 0.84$ between the two metrics. This confirms that \textbf{oMeBench} conclusions are robust across both functional-group-centered and graph-structured similarity measures.

We retain Tanimoto as the primary metric of \textbf{oMeS-partial} because mechanistic reasoning is fundamentally about tracking stepwise bond formation and cleavage, which Tanimoto captures more directly. GESim, while useful for assessing scaffold consistency, may over-credit predictions that preserve global topology but miss the key local electronic changes that define a mechanism.

\begin{table}[!htb]
\centering
\small
\begin{tabular}{lcc}
\toprule
\textbf{Model} & $S_{part}$ \textbf{(Tanimoto)} & $S_{part}$ \textbf{(GESim)} \\
\midrule
Gemini-Pro-2.5   & 37.87 & 50.41 \\
GPT-5            & 29.11 & 45.62 \\
DeepSeek-R1      & 28.40 & 45.53 \\
o3               & 28.11 & 42.42 \\
Claude-Sonnet-4  & 17.92 & 35.11 \\
\midrule
GPT-OSS          & 14.19 & 24.81 \\
GPT-4o           & 5.39  & 15.19 \\
Qwen-3           & 4.27  & 9.92  \\
LLaMA-3          & 3.27  & 8.19  \\
Mistral          & 2.25  & 7.16  \\
\midrule
OLMo-2           & 0.56  & 1.25  \\
OpenBioLLM       & 0.29  & 0.64  \\
ChemDFM          & 0.00  & 0.24  \\
\bottomrule
\end{tabular}
\caption{$S_{part}$ scores computed with Tanimoto (Morgan fingerprints) vs.\ GESim. Spearman correlation between the two: $\rho = 0.84$.}
\label{tab:similarity_ablation}
\end{table}

\newpage
\section{Alignment Strategy}
\label{app:alignment}
The alignment strategy is a specialized dynamic programming (DP) algorithm designed to align predicted reaction mechanism steps ($\mathbf{P}$) against a set of gold-standard steps ($\mathbf{G}$). It significantly improves upon Minimum Edit Distance (MED) and simple match-prioritized scoring by incorporating \textbf{chemical structural similarity} and a \textbf{hierarchical preference for alignment} into its scoring function.

\subsection{Alignment Process and Multi-Criteria Scoring}
The alignment strategy is shown as the below,
\begin{tcolorbox}[title=\textbf{Pseudocode for Alignment Strategy}, colback=white, colframe=black, fonttitle=\bfseries, colbacktitle=gray!15, boxrule=0.6pt, arc=3mm]

\textbf{Input:} 
- Gold steps $G = [(type, smiles, weight)]_{1..N}$ 
- Predicted steps $P = [(type, smiles)]_{1..M}$ 
- Similarity function $sim(\cdot,\cdot)$, threshold $\tau$

\textbf{Output:} oMeSResult $(S_{tot}, S_{part}, Alignment, V, L)$

\begin{algorithmic}[1]
\State Initialize DP tables: $DP\_tot, DP\_par, DP\_pen, DP\_rank, Trace$
\State Fill first row/col with skip penalties (skip\_gold / skip\_pred)

\For{$i=1 \to N$}
  \For{$j=1 \to M$}
    \State Extract $g=(g\_type, g\_smi, w)$, $p=(p\_type, p\_smi)$
    \If{$g\_type = p\_type$}
       \State $m\_tot \gets w$ if canonical($g\_smi$) = canonical($p\_smi$)
       \State $m\_par \gets w \cdot \sigma(sim(g,p), \tau)$
       \State action $\gets$ ``match"
    \Else
       \State $m\_tot, m\_par \gets 0$
       \State action $\gets$ ``type\_mismatch"
    \EndIf
    \State Candidates $\gets$ 
       \begin{itemize}
         \item Skip\_gold: from $(i-1,j)$
         \item Skip\_pred: from $(i,j-1)$
         \item Diagonal: from $(i-1,j-1)$ with $(m\_tot, m\_par)$
       \end{itemize}
    \State Choose best candidate by priority: 
       \[
         \max(\;DP\_tot, DP\_par, Rank, DP\_pen\;)
       \]
    \State Update DP and Trace at $(i,j)$
  \EndFor
\EndFor

\Statex

\textbf{Backtrace:}
\State Start from $(N,M)$, follow Trace until $(0,0)$
\State Build Alignment sequence (match / mismatch / skip)

\Statex

\textbf{Metrics:}
\State $S_{tot} \gets DP\_tot[N][M]$
\State $S_{part} \gets DP\_par[N][M]$
\State $V \gets$ fraction of valid predicted SMILES
\State $L \gets$ fraction of matches in alignment

\State \Return $(S_{tot}, S_{part}, Alignment, V, L)$
\end{algorithmic}

\end{tcolorbox}

Unlike traditional sequence alignment, which maximizes a single score or minimizes a single cost, $oMeS$ uses a \textbf{four-dimensional, lexicographically ordered score vector} at each DP cell $\text{DP}[i][j]$ to determine the optimal path:

$$\text{Key} = (S_{\text{total}}, S_{\text{partial}}, \text{Rank}, S_{\text{penalty}})$$

The optimal action (Match, Type Mismatch, Skip Gold, or Skip Pred) is chosen by maximizing this vector in the order presented (i.e., $S_{\text{total}}$ is the primary objective, followed by $S_{\text{partial}}$, etc.). The four fundamental actions used in the DP transitions are described in Table~\ref{tab:actions}.

\begin{table}[h!]
\centering
\begin{tabular}{@{}ll@{}}
\toprule
\textbf{Action} & \textbf{Description} \\
\midrule
\texttt{skip\_gold} & A step present in the \textbf{Gold} sequence was skipped by the \textbf{Pred} sequence. \\
\texttt{skip\_pred} & A step present in the \textbf{Pred} sequence is extraneous and does not match any \textbf{Gold} step. \\
\texttt{match} & The steps align and correctly choose the step types \\
\texttt{type\_mismatch} & The steps align but choose the wrong step types. \\
\bottomrule
\end{tabular}
\caption{Description of Alignment Actions}
\label{tab:actions}
\end{table}

\begin{itemize}
    \item $S_{\text{total}}$: \textbf{Strict Match Score (1st Priority):} Rewards perfect alignment where the step \textbf{type} is identical and the \textbf{canonical SMILES} are identical. This is the highest reward.
    \item $S_{\text{part}}$: \textbf{Partial Chemical Match Score (2nd Priority):} When $S_{\text{total}}=0$, this metric rewards alignment based on \textbf{chemical structural similarity}. It is calculated using the Tanimoto similarity of Morgan fingerprints, but only if the similarity is greater than a threshold $\tau=0.60$ (Eq.~\ref{eq:step-sigma}).
    \begin{equation}
        \label{eq:step-sigma}
        \sigma(s_g, s_p) = \begin{cases} \text{Tanimoto}(s_g, s_p) & \text{if } \text{Tanimoto}(s_g, s_p) \ge \tau \\ 0.0 & \text{otherwise} \end{cases}
    \end{equation}
    \item Rank: \textbf{Explicit Operation Priority (3rd Priority):} This ensures the DP prioritizes alignment over skipping when $S_{\text{total}}$ and $S_{\text{partial}}$ are equal.

The ranks are defined as:
    \begin{center}
        \begin{tabular}{ll}
            \toprule
            Action & Rank \\
            \midrule
            $\text{match}$ & 3 \\
            $\text{type\_mismatch}$ & 2 \\
            $\text{skip\_gold}$ / $\text{skip\_pred}$ & 1 \\
            \bottomrule
        \end{tabular}
    \end{center}
    \item $S_{\text{penalty}}$: Non-Match Penalty (4th Priority): A minor penalty ($-1\mathrm{e}{-6}$) applied to skip and type\_mismatch actions. This minimizes the total number of non-perfect alignments when all higher-priority scores are identical.
\end{itemize}

\subsection{Edge Cases Demonstrating Robustness of the Alignment Strategy}
\label{app:edge_cases_final}

We denote each gold mechanistic step as $(T_i, I_i)$, where $T_i$ represents the reaction type and $I_i$ the intermediate.  
Predicted steps are $(T'_j, I'_j)$.  
Each alignment result is displayed in two rows:  
the upper row shows the predicted types $T'$, and the lower row shows the alignment label (\textit{Match}, \textit{Type Mismatch}, \textit{Skip\_gold}, \textit{Skip\_pred}).  
Columns labeled “/” indicate redundant predictions that do not correspond to any gold step.

\paragraph{Edge Case 1: Long but redundant predictions.}
In this situation, the model outputs a long sequence containing extra but chemically valid substeps, such as protonation–deprotonation or intermediate resonance rearrangements.  
While these steps are redundant, they remain chemically reasonable.  
Traditional MED or match-first DP tends to greedily match these extra predictions, misplacing true gold correspondences and therefore underestimating the model's performance.  
Our alignment strategy correctly preserves structural consistency, skipping only redundant steps (\texttt{Skip\_pred}) without penalizing chemically correct reasoning.

\vspace{-1mm}
\begin{center}
\resizebox{\textwidth}{!}{
\begin{tabular}{@{}lccccccc@{}}
\toprule
\textbf{Gold} 
& / 
& $(T_1{=}\mathrm{A}, I_1{=}S_1)$ 
& / 
& / 
& $(T_2{=}\mathrm{B}, I_2{=}S_2)$ 
& $(T_3{=}\mathrm{C}, I_3{=}S_3)$ 
& $(T_4{=}\mathrm{D}, I_4{=}S_4)$ \\
\midrule
\textbf{Our Alignment} 
& A & A & A & - & B & C & D \\
& Skip\_pred & Match & Skip\_pred & - & Match & Match & Match \\
\addlinespace[2pt]
\textbf{MED / match-first DP} 
& - & A & A & A & B & C & D \\
& - & Match & Skip\_pred & Skip\_pred & Match & Match & Match \\
\bottomrule
\end{tabular}
}
\end{center}

\vspace{1mm}
In this case, our strategy recognizes that the additional $(T'_1, I'_1)$ and $(T'_3, I'_3)$ are redundant but not erroneous, yielding an accurate evaluation of mechanistic reasoning.  
By contrast, conventional DP misaligns these intermediate steps, lowering both $S_\text{total}$ and $S_\text{partial}$ despite chemically valid logic.

\paragraph{Edge Case 2: Incomplete but near-correct predictions.}
Here, the model predicts a shorter sequence that captures most intermediates correctly but misclassifies the \textbf{final step type}.  
For instance, it correctly generates $(C,S_3)$ and $(C,S_5)$ but predicts $(E,S_6)$ instead of the correct $(D,S_6)$.  
This single type misclassification leads to a cascade in traditional DP: the alignment path “fetches” the wrong step, propagating mismatches backward and producing multiple \texttt{Skip\_gold} actions.  
Our hierarchical scoring isolates this terminal type error without disturbing earlier matches, thus preserving earned partial credit.

\vspace{-1mm}
\begin{center}
\resizebox{\textwidth}{!}{
\begin{tabular}{@{}lcccccc@{}}
\toprule
\textbf{Gold} 
& $(A,S_1)$ & $(B,S_2)$ & $(C,S_3)$ & $(E,S_4)$ & $(C,S_5)$ & $(D,S_6)$ \\
\midrule
\textbf{Our Alignment} 
& A & C & - & - & C & E \\
& Match & Match & Skip\_gold & Skip\_gold & Match & Type Mismatch \\
\addlinespace[2pt]
\textbf{MED DP} 
& A & C & C & E & - & - \\
& Match & Match & Match & Type Mismatch & Skip\_gold & Skip\_gold \\
\addlinespace[2pt]
\textbf{Match-first DP} 
& A & C & C & E & - & - \\
& Match & Match & Match & Type Mismatch & Skip\_gold & Skip\_gold \\
\bottomrule
\end{tabular}
}
\end{center}

\vspace{1mm}
This example highlights that even a minor terminal error (incorrect type $(E,S_6)$) can distort traditional DP scoring, leading to overcounted skips and underestimation of reasoning quality.  
Our approach, however, penalizes only the actual type mismatch while retaining the credit for structurally and sequentially correct steps.

\paragraph{Takeaways.}
Across both cases, our alignment method:
\begin{itemize}
    \item Accurately distinguishes between \textbf{chemically redundant} and \textbf{truly erroneous} steps, preventing underestimation of valid reasoning.
    \item Localizes \textbf{late-step classification errors}, avoiding backward propagation and preserving consistent credit distribution.
\end{itemize}
ensuring more faithful measurement of multi-step mechanistic reasoning accuracy.

\newpage

\section{Extracting High-Similarity Reactions via DRFP}
\label{sec:reaction-retrieval}

To provide contextually relevant examples for in-context learning or to augment reaction datasets, we develop a retrieval mechanism that identifies structurally similar reactions from an external chemical reaction database. Our approach leverages DRFP to measure similarity between chemical transformations, enabling efficient retrieval of the most analogous reactions.

\subsection{Reaction Representation}

Given a chemical reaction with reactants $\{R_1, R_2, \ldots, R_m\}$ and products $\{P_1, P_2, \ldots, P_n\}$ represented as SMILES strings, we first construct a unified reaction SMILES representation:
\begin{equation}
\text{RXN}_{\text{SMILES}} = \underbrace{R_1.R_2.\cdots.R_m}_{\text{reactants}} \gg \underbrace{P_1.P_2.\cdots.P_n}_{\text{products}}
\end{equation}
where ``$.$'' concatenates multiple molecules and ``$\gg$'' separates reactants from products.

\subsection{DRFP}

Traditional molecular fingerprints encode individual molecules as binary or count vectors based on structural features. For reactions, we employ the DRFP approach~\citep{rdkit}, which captures the structural changes occurring during a chemical transformation.

For each structural feature $k \in \mathcal{K}$ (e.g., circular substructures of varying radii derived from Morgan fingerprints~\citep{rogers2010extended}), we compute the feature count in reactants and products:
\begin{align}
c_k^{(R)} &= \sum_{i=1}^{m} f_k(R_i), \\
c_k^{(P)} &= \sum_{j=1}^{n} f_k(P_j),
\end{align}
where $f_k(M)$ denotes the count of feature $k$ in molecule $M$. The difference fingerprint is then defined as:
\begin{equation}
\mathbf{d} = (d_k)_{k \in \mathcal{K}}, \quad \text{where} \quad d_k = c_k^{(P)} - c_k^{(R)}.
\end{equation}

A positive value $d_k > 0$ indicates that feature $k$ is generated during the reaction, while $d_k < 0$ indicates consumption. This signed integer vector encodes the net structural changes, making it particularly suitable for comparing reaction mechanisms.

In practice, we use RDKit's \texttt{CreateDifferenceFingerprintForReaction} function, which internally hashes the feature space $\mathcal{K}$ into a fixed-dimensional sparse vector via:
\begin{equation}
b = h(k) \bmod N,
\end{equation}
where $h(\cdot)$ is a hash function and $N$ is the fingerprint dimension. Feature counts are accumulated into corresponding hash buckets, yielding a compact representation.

\subsection{Similarity Computation and Retrieval}

Given a query reaction with fingerprint $\mathbf{d}_q$ and a candidate reaction with fingerprint $\mathbf{d}_c$ from an external database, we compute their Tanimoto similarity:
\begin{equation}
\text{Sim}(\mathbf{d}_q, \mathbf{d}_c) = \frac{\sum_{k} \min(|d_{q,k}|, |d_{c,k}|)}{\sum_{k} \max(|d_{q,k}|, |d_{c,k}|)},
\end{equation}
where the absolute values ensure that similarity focuses on the magnitude of structural changes regardless of sign. RDKit's \texttt{TanimotoSimilarity} function handles this computation efficiently for sparse count vectors.

Our retrieval algorithm proceeds as follows:
\begin{enumerate}
    \item \textbf{Database preprocessing}: We pre-compute difference fingerprints for all reactions in the external database and cache them as $\{(\mathbf{d}_i, \text{reaction}_i)\}_{i=1}^{|\mathcal{D}|}$.
    \item \textbf{Query fingerprint generation}: For a query reaction, we compute its difference fingerprint $\mathbf{d}_q$ using the procedure described above.
    \item \textbf{Similarity scoring}: We compute Tanimoto similarity between $\mathbf{d}_q$ and each cached fingerprint $\mathbf{d}_i$, filtering out identical reactions (by reaction ID or exact SMILES match).
    \item \textbf{Top-$k$ selection}: We rank candidates by decreasing similarity and retrieve the top $k$ (typically $k=3$) most similar reactions.
\end{enumerate}

\subsection{Implementation Details}

We implement this retrieval system using RDKit version 2023.09 with Python 3.9. The external database consists of $N_{\text{ext}}$ standardized reactions from diverse sources, stored in JSONL format with fields for reactant SMILES, product SMILES, and reaction metadata (conditions, mechanisms, etc.). The difference fingerprints are computed using Morgan-based circular environments with default RDKit parameters. To avoid computational redundancy, we cache all database fingerprints in memory upon first access.

This fingerprint-based retrieval enables efficient identification of mechanistically similar reactions without requiring explicit reaction classification or rule-based matching, making it well-suited for augmenting language model inputs with relevant chemical knowledge.

\newpage

\section{Prompts}
\subsection{Prompt for suggested group}
\label{app:prompt_suggest_groups}
\begin{tcolorbox}[title={\textbf{Prompt Used for R-group Suggestions}}, colback=gray!5, colframe=black!40, fonttitle=\bfseries]
You are an expert organic chemist. I need \texttt{\{num\_suggestions\}} chemically reasonable substituents for each R-group placeholder.

\vspace{0.5em}
\noindent\textbf{Reaction Template:} \texttt{\{template\}} \\
\textbf{R-groups to substitute:} \texttt{\{', '.join(rgroups)\}}

\vspace{0.5em}
\noindent\textbf{Requirements:}
\begin{enumerate}
    \item Provide valid SMILES strings (no placeholder notation)
    \item Include diverse functional groups: alkyl, aryl, heteroatoms, etc.
    \item Do not include any H (H, CH\textsubscript{3}, CH\textsubscript{2}CH\textsubscript{3}, etc.) in the suggestions, only include C instead (C, CC)
    \item Ensure chemical stability and synthetic accessibility
    \item Include both electron-donating and electron-withdrawing groups
    \item If there are already ring systems in the template, do not include any ring systems in the suggestions
\end{enumerate}

\vspace{0.5em}
\noindent\textbf{Format your response as JSON, below is an example}
\begin{verbatim}
{
    "[*:1]": ["C", "CC", "C(C)(C)C", "c2ccccc2", "CCCC(C)C", 
              "CC(C)(C)OC1=CC=C(C=C1)C2=CC=C(OC(C)(C)C)C=C2", 
              "C(F)(F)F"],
    "[*:2]": ["C", "c2ccccc2", "CCCCCCCCCC4=CC2=CC=CC=C2C3=CC=CC=C34", 
              "COC1=CC(=CC=C1)C2=CC(=CC=C2O[Si](C)(C)C)C3=CC=CC=C3", 
              "OC", "CC(C)(C)C1=CC(OCCC(C)C)=CC=C1"]
}
\end{verbatim}

\vspace{0.5em}
\noindent\textbf{Constraint:} Only return the JSON, no other text.
\end{tcolorbox}

\newpage

\subsection{Prompt used for rationale annoation}
\label{app:prompt_rationale}
Chain-of-Thought SFT (CoT-SFT) data is used to train Qwen-3-4B in Section~\ref{sec:improving}. Compared with standard SFT, CoT-SFT additionally includes a natural-language rationale for each mechanism step. The two training targets can be abstracted as:
\begin{itemize}
    \item \textbf{SFT:} \texttt{\{step, type, subtype, intermediate\}}
    \item \textbf{CoT-SFT:} \texttt{\{step, rationale, type, subtype, intermediate\}}
\end{itemize}
The rationales are generated by Gemini-2.5-Pro using the prompt shown below.

\begin{tcolorbox}[title={\textbf{Prompt used for rationale annoation}}, colback=gray!5, colframe=black!40, fonttitle=\bfseries]
You are an organic chemist. Given a reaction JSON (reactants, products, conditions) and a step-by-step mechanism list without explanations, your task is to:

\begin{itemize}
    \item Add a top-level \texttt{"description"} (overall reasoning, maximum 90 words),
    \item Add a \texttt{"rationale"} (maximum 40 words) to each mechanism step.
\end{itemize}

You should think like a chemist, and for each step, explain \textbf{why} it happens under the given conditions, not just what happens.

\vspace{0.5em}
\noindent\textbf{Output Format:}
\begin{verbatim}
{
  "description": "overall reasoning",
  "mechanism": [
    {"step": 1, "rationale": "rationale1"},
    {"step": 2, "rationale": "rationale2"},
    ...
  ]
}
\end{verbatim}

\vspace{0.5em}
\noindent\textbf{Hints:} In each step's rationale, include at least one \textit{specific chemical reasoning element}, such as:
\begin{itemize}
    \item Electronic effects: resonance stabilization, inductive effects, conjugation, electron-withdrawing/donating groups.
    \item Steric effects: bulky substituents blocking attack.
    \item Reaction conditions: acid/base catalysis, solvent polarity, temperature effects.
    \item Kinetics vs.\ thermodynamics: faster pathway vs.\ more stable product.
    \item Leaving group ability and stability of intermediates.
    \item Structure analysis: how the intermediate structure approaches the product.
\end{itemize}

\vspace{0.5em}
\noindent\textbf{Key Constraints:}
\begin{itemize}
    \item Refer to placeholders as \texttt{R1 ([*:1])}, \texttt{R2 ([*:2])}, \texttt{R3 ([*:3])}; discuss sterics/electronics qualitatively only.
    \item \textbf{Do not} add, remove, or reorder steps; the number should exactly match the original mechanism list.
    \item Output \textbf{valid JSON only}, and do \textbf{not} include any extra text.
\end{itemize}
\end{tcolorbox}

Note that the provided hints are designed to follow the logic that human chemists typically adopt when analyzing mechanisms, essentially serving as a chemically grounded "mind map". This design has proven effective in guiding models to produce results with more meaningful underlying chemical analysis.

\subsection{Prompt used for mechanism generation}
\label{app:prompt_default}
\begin{tcolorbox}[title={\textbf{Prompt Used for Mechanism Generation}}, colback=gray!5, colframe=black!40, fonttitle=\bfseries]
You are an expert in organic reaction mechanisms.

\vspace{0.5em}
\noindent\textbf{Task:}  

Given the following information:\textbf{Reactants (SMILES)},\textbf{Products (SMILES)}, \textbf{Reaction Conditions}, your task is to write a \textbf{step-by-step} organic chemistry reaction mechanism as a \textbf{JSON list}.

\vspace{1em}
\noindent\textbf{For each step, include:}
\begin{enumerate}
    \item \texttt{``step"} - the step number (starting from 1)
    \item \texttt{``type"} - the reaction type, selected \textbf{exactly} from the predefined list below
    \item \texttt{``subtype"} - the reaction subtype, selected \textbf{exactly} from the predefined list below
    \item \texttt{``intermediate\_smiles"} - a valid SMILES string representing the intermediate at this step (must be parsable by RDKit)
\end{enumerate}

\vspace{0.5em}
\noindent\textbf{Predefined list of types and subtypes:}
\begin{verbatim}
[
  {"type": "...", "subtype": "...", "description":"..."},
  ... (full content provided in Table 6)
]
\end{verbatim}

\vspace{0.5em}
\noindent\textbf{Output format:}  

Return only the \textbf{valid JSON} in the format below. Do not include any explanatory text, code blocks, markdown formatting, or any other content before or after the JSON.
\begin{verbatim}
[
  {{ "step": 1, "type": "", "subtype": "", "intermediate_smiles": "" }},
  ...
]
\end{verbatim}

\vspace{0.5em}
\noindent\textbf{Example:}
\begin{verbatim}
Input: # EXAMPLE INPUT OMITTED FOR BREVITY
Expected Output: # EXAMPLE OUTPUT OMITTED FOR BREVITY

\end{verbatim}

\vspace{0.5em}
\noindent\textbf{Input for you:}
\begin{verbatim}
[
  {
    "reactants_smiles": "{ reactants_smiles }", 
    "products_smiles": "{ products_smiles }",
    "conditions": "{ conditions }"
  }
]
\end{verbatim}

\noindent\textbf{Now, generate your output!}
\end{tcolorbox}

\clearpage
\newpage

\section{Error Discussions}

\begin{table*}[!htb]
\centering
\resizebox{\textwidth}{!}{
\begin{tabular}{p{3.2cm} p{4cm} p{5.5cm} p{5.5cm}}
\toprule
\textbf{Category} & \textbf{Subtype} & \textbf{Manifestation} & \textbf{Cause Analysis} \\
\midrule
\textbf{Chemical Knowledge Errors}
& Reaction condition misinterpretation & 
Under acidic conditions, the model performs nucleophilic attack or basic deprotonation; under transition-metal catalysis, fails to include coordination or oxidative addition steps. & 
Models may treat reaction conditions (``acid", ``base") as plain text tokens rather than mechanistically decisive entities. Training data often states that a reaction ``occurs under acidic conditions" but lacks stepwise mechanistic detail linking conditions to reaction pathways. \\
\cmidrule(lr){2-4}
& Electrophilic/nucleophilic center misidentification & 
Nucleophiles attack incorrect atoms or positions; conjugated systems attacked at non-electron-rich sites (e.g., $\beta$-position rather than carbonyl carbon). & 
Deficiency in reasoning about electronic effects (inductive, conjugative, steric). Models may rely only on functional group ``canonical" reactivity rather than context-specific micro-environment. \\
\cmidrule(lr){2-4}
& Valence/charge rule violation & 
Generated intermediates contain pentavalent carbon, tetravalent nitrogen (non-ammonium), or trivalent oxygen (non-oxonium); misplaced or imbalanced charges. & 
Represents fundamental violation of chemical rules. As sequence predictors, LLMs lack an internal chemical engine to enforce valence/electron conservation, often producing illogical SMILES patterns. \\
\cmidrule(lr){2-4}
& Structural/steric constraint neglect & 
Failure to enforce anti-periplanar geometry in E2/SN2 reactions; impossible pathways in cyclic systems due to steric hindrance. & 
Stems from difficulty in inferring 3D spatial constraints from 1D/2D textual representations (e.g., SMILES). \\
\cmidrule(lr){2-4}
& Intermediate stability misjudgment & 
SN1/E1 mechanisms without carbocation rearrangement; generation of unstable intermediates without subsequent rapid conversion. & 
Model fails to assess relative energies of intermediates; rearrangements require non-trivial reasoning beyond pattern matching. \\
\midrule
\textbf{Mechanistic and Procedural Errors}
& Mechanism
misclassification & 
Misclassifies nucleophilic substitution as electrophilic substitution; describes radical reactions using ionic steps; decomposes concerted pericyclic reactions into stepwise ionic events. & 
Failure to recognize fundamental reaction type (e.g., light-initiated radical vs.\ ionic pathways). \\
\cmidrule(lr){2-4}
& Step omission or illogical jumps & 
Missing key rearrangements; skipping intermediates (e.g., jumping from nucleophilic attack directly to product without deprotonation). & 
Models may have been exposed to simplified mechanistic descriptions or aim to ``optimize" step count, reflecting insufficient grasp of mechanistic rigor. \\
\midrule
\textbf{Structural and Formalization Errors}
& Invalid SMILES representation & 
Generated SMILES cannot be parsed (e.g., by RDKit) or parse into chemically nonsensical structures (e.g., open-chain rings). & 
LLMs struggle with precise symbolic sequences; minor character errors (brackets, ring indices) break validity. \\
\cmidrule(lr){2-4}
& JSON/field mismatches & 
Outputs not valid JSON; \texttt{type}/\texttt{subtype} contain out-of-vocabulary strings; missing required fields. & 
Reflects instruction-following limitations. Models may lose adherence to schema constraints over long generations. \\
\bottomrule
\end{tabular}
}
\caption{Taxonomy of error types observed in model-generated mechanisms. Errors are grouped into three broad categories, each with representative subtypes, manifestations, and cause analysis.}
\label{tab:error_taxonomy_app}
\end{table*}

\newpage

\section{Case Studies}
\label{app:case-study}

\subsection{Case Study 1}
\label{app:case-study-1}

Reactants: \texttt{C(N)C(=O)c1ccccc1, CCOC(=O)CC(=O)C} \\
Products: \texttt{CCOC(=O)c1c(C)Ncc1c2ccccc2} \\
Conditions: Acidic medium

\begin{table}[htbp]
\centering
\small
\begin{tabular}{p{0.06\textwidth} p{0.55\textwidth} p{0.32\textwidth}}
\toprule
\textbf{Step} & \textbf{Model-Generated Mechanism} & \textbf{Expert Comment} \\
\midrule
1 & \textbf{Type}: addition / \textbf{Subtype}: nucleophilic\_addition \newline
    \textbf{Intermediate}: \texttt{\detokenize{CCOC(=O)C(C(=O)C)C(O)(c1ccccc1)CN}} 
    & Completely illogical. Both the nucleophilic and electrophilic sites are incorrect, leading to a random adduct that bears no mechanistic connection to the product. \\
\hline
2 & \textbf{Type}: elimination / \textbf{Subtype}: leaving\_group\_elimination \newline
    \textbf{Intermediate}: \texttt{\detokenize{CCOC(=O)C(=C(c1ccccc1)CN)C(=O)C}} 
    & Structural errors persist. While the reaction type (elimination) is plausible, the intermediate structure diverges significantly from the correct mechanistic pathway. \\
\hline
3 & \textbf{Type}: addition / \textbf{Subtype}: nucleophilic\_addition \newline
    \textbf{Intermediate}: \texttt{\detokenize{CCOC(=O)C1=C(c2ccccc2)CNC1(C)O}} 
    & Mechanistic reasoning error. The model incorrectly assumes nitrogen participates as a nucleophile again, whereas the reaction should involve the enamine carbon center. \\
\hline
4 & \textbf{Type}: elimination / \textbf{Subtype}: leaving\_group\_elimination \newline
    \textbf{Intermediate}: \texttt{\detokenize{CCOC(=O)C1=C(c2ccccc2)C=NC(C)=C1}} 
    & Ring system error. The model inexplicably transforms the five-membered ring into a six-membered one, which is chemically impossible under the given conditions. \\
\hline
5 & \textbf{Type}: proton\_transfer / \textbf{Subtype}: acid\_base\_proton\_transfer \newline
    \textbf{Intermediate}: \texttt{\detokenize{CCOC(=O)C1=C(C)NC(c2ccccc2)=CC1}} 
    & The same ring-size error continues. The overall sequence fails to restore the correct conjugation pattern. \\
\hline
6 & \textbf{Type}: elimination / \textbf{Subtype}: leaving\_group\_elimination \newline
    \textbf{Intermediate}: \texttt{\detokenize{CCOC(=O)c1c(C)N=C(c2ccccc2)C=C1}} 
    & Although the final structure matches the product, the mechanistic path remains chemically unsound, indicating weak self-correction ability of the model. \\
\bottomrule
\end{tabular}
\caption{Model-generated mechanism for Case Study 1.}
\end{table}

\paragraph{Error Analysis.}
While large language models exhibit clear advantages in handling simple ionic reactions with relatively accurate stepwise reasoning, they remain limited in representing Lewis acid catalysis and complex ring-forming systems.  
In radical or catalytic cycle reactions, the model struggles to construct a coherent, chemically valid pathway and tends to generate only substrate–product mappings.  
Moreover, persistent issues include misplaced protonation timing, loss of hydrogen atoms, and poor control of stereochemical details-all of which highlight the current gap between linguistic coherence and genuine chemical reasoning.

\newpage

\subsection{Case Study 2}
\label{app:case-study-2}
 
Reactants: \texttt{COc1cc(Br)c(OC)cc1, COC(Cl)Cl, Cl[Ti](Cl)(Cl)Cl, Cl[Ti](Cl)(Cl)Cl} \\
Products: \texttt{COc1cc(Br)c(OC)cc1(C=O)} \\
Conditions: 1. Lewis acid; 2. Aqueous Hydrolysis

\begin{table}[htbp]
\centering
\small
\begin{tabular}{p{0.06\textwidth} p{0.55\textwidth} p{0.32\textwidth}}
\toprule
\textbf{Step} & \textbf{Model-Generated Mechanism} & \textbf{Expert Comment} \\
\midrule
1 & \textbf{Type}: heterolytic\_cleavage \newline
    \textbf{Intermediate}: \texttt{\detokenize{CO[C+]Cl}} & Generated Chemically Invalid SMILES. The model doesn't realize that the hydrogen atom should be put into the middle brackets to form a correct SMILES structure. This is a mistake that the model frequently make here and there. \\
\hline
2 & \textbf{Type}: electrophilic\_substitution \newline
    \textbf{Intermediate}: \texttt{\detokenize{COc1c(OC)c(Br)c(C(Cl)OC)c[cH+]1}} & Incorrect Structure. The model doesn't realize the fact that the electrophilic substitution broke the aromaticity of the ring, thus  resulting in the lowercase spelling of benzene ring to be  incorrect. \\
\hline
3 & \textbf{Type}: acid\_base\_proton\_transfer \newline
    \textbf{Intermediate}: \texttt{\detokenize{COc1c(OC)c(Br)c(C(Cl)OC)cc1}} & Partly Incorrect Step Logic. The proton transfer after an electrophilic addition in an aromatic ring is more often regarded as proton elimination. \\
\hline
4 & \textbf{Type}: heterolytic\_cleavage \newline
    \textbf{Intermediate}: \texttt{\detokenize{COc1c(OC)c(Br)c(C=[O+]C)cc1}} & Correct \\
\hline
5 & \textbf{Type}: nucleophilic\_addition \newline
    \textbf{Intermediate}: \texttt{\detokenize{COc1c(OC)c(Br)c(C(OC)[OH2+])cc1}} & Correct \\
\hline
6 & \textbf{Type}: acid\_base\_proton\_transfer \newline
    \textbf{Intermediate}: \texttt{\detokenize{COc1c(OC)c(Br)c(C(O)OC)cc1}} & Correct \\
\hline
7 & \textbf{Type}: acid\_base\_proton\_transfer \newline
    \textbf{Intermediate}: \texttt{\detokenize{COc1c(OC)c(Br)c(C(O)[O+](C)H)cc1}} & Generated Chemically Invalid SMILES. The order of carbon and hydrogen atoms are incorrect. \\
\hline
8 & \textbf{Type}: leaving\_group\_elimination \newline
    \textbf{Intermediate}: \texttt{\detokenize{COc1c(OC)c(Br)c(C=[OH+])cc1}} & Correct \\
\hline
9 & \textbf{Type}: acid\_base\_proton\_transfer \newline
    \textbf{Intermediate}: \texttt{\detokenize{COc1c(OC)c(Br)c(C=O)cc1}} & Correct \\
\bottomrule
\end{tabular}
\caption{Model-generated mechanism for Case Study 2.}
\end{table}

\newpage
\subsection{Case Study 3}
\label{app:case-study-3}

Reactants: \texttt{Oc1ccccc1, CC\#N} \\
Products: \texttt{CC(=O)c1ccccc1O} \\
Conditions: i. ZnCl$_2$, HCl; ii. H$_2$O

\begin{table}[htbp]
\centering
\small
\begin{tabular}{p{0.06\textwidth} p{0.55\textwidth} p{0.32\textwidth}}
\toprule
\textbf{Step} & \textbf{Model-Generated Mechanism} & \textbf{Expert Comment} \\
\midrule
1 & \textbf{Type}: acid\_base\_proton\_transfer \newline
    \textbf{Intermediate}: \texttt{\detokenize{CC#[NH+]}} & Incorrect Catalysis. The reaction should be initiated by Lewis acid activation, but the model incorrectly uses a proton as the catalyst. This is a recurring error where the model often ignores the role of Lewis acids. \\
\hline
2 & \textbf{Type}: electrophilic\_substitution \newline
    \textbf{Intermediate}: \texttt{\detokenize{Oc1c(C(=[NH+])C)C[cH+]cc1}} & Incorrect Structure. The position of the double bond is wrong. The model fails to capture that the hydroxyl group donates electron density through conjugation, which makes the aromatic ring nucleophilic. \\
\hline
3 & \textbf{Type}: acid\_base\_proton\_transfer \newline
    \textbf{Intermediate}: \texttt{\detokenize{CC(=[NH+])c1ccccc1O}} & Incorrect SMILES. In addition to misrepresenting the Lewis acid as a proton, the SMILES structure is invalid as it places an unpaired electron on nitrogen, which is chemically unreasonable. \\
\hline
4 & \textbf{Type}: nucleophilic\_addition \newline
    \textbf{Intermediate}: \texttt{\detokenize{CC(O[H2+])c1ccccc1O}} & Incorrect SMILES. The nitrogen atom is again assigned an unpaired electron, reflecting a repeated formatting error. \\
\hline
5 & \textbf{Type}: acid\_base\_proton\_transfer \newline
    \textbf{Intermediate}: \texttt{\detokenize{CC(O)c1ccccc1O}} & Incorrect SMILES. The same issue persists with the misrepresentation of charged nitrogen species. \\
\hline
6 & \textbf{Type}: acid\_base\_proton\_transfer \newline
    \textbf{Intermediate}: \texttt{\detokenize{C(O)c1ccccc1O}} & Incorrect SMILES. Once again, the model incorrectly places an unpaired electron on nitrogen. \\
\hline
7 & \textbf{Type}: acid\_base\_proton\_transfer \newline
    \textbf{Intermediate}: \texttt{\detokenize{CC(=[OH+])c1ccccc1O}} & Correct. \\
\hline
8 & \textbf{Type}: acid\_base\_proton\_transfer \newline
    \textbf{Intermediate}: \texttt{\detokenize{CC(C=O)c1ccccc1O}} & Correct. \\
\bottomrule
\end{tabular}
\caption{Model-generated mechanism for Case Study 3.}
\end{table}

\newpage

\subsection{Case Study 4}
\label{app:case-study-4}

\textbf{Reaction Overview.}  
\textbf{Reactants:} \texttt{CC(C1CC1)=NNS(=O)(=O)c2ccc(C)cc2} \\
\textbf{Products:} \texttt{C=CC1CC1} \\
\textbf{Conditions:} Base

\begin{table}[htbp]
\centering
\resizebox{\textwidth}{!}{
\begin{tabular}{p{0.04\textwidth} p{0.48\textwidth} p{0.43\textwidth}}
\toprule
\textbf{Step} & \textbf{Model-Generated Mechanism} & \textbf{Expert Comment} \\
\midrule
1 & \textbf{Type:} acid\_base\_proton\_transfer \newline
    \textbf{Intermediate:} \texttt{CC(C1CC1)=N[N-]S(=O)(=O)\newline c1ccc(C)cc1} &
    Correct deprotonation of the hydrazone nitrogen. This step aligns with the gold mechanism. \\
\hline

2 & \textbf{Type:} acid\_base\_proton\_transfer \newline
    \textbf{Intermediate:} \texttt{[CH2-]C(C1CC1)=N[N-]S(=O)(=O)\newline
    c1ccc(C)cc1} &
    Incorrect second deprotonation: removes an $\alpha$-hydrogen instead of initiating N--N bond cleavage. Mechanistically implausible. \\
\hline
3 & \textbf{Type:} leaving\_group\_elimination \newline
    \textbf{Intermediate:} \texttt{C=C(C1CC1)N=N} &
    Partial match. Correctly identifies diazo formation but misses the ionic character (\texttt{[N+]=[N-]}). Electron flow and charge distribution are wrong. \\
\hline
4 & \textbf{Type:} leaving\_group\_elimination \newline
    \textbf{Intermediate:} \texttt{C=[C-]C1CC1} &
    Incorrect elimination: the carbocation intermediate is replaced by a carbanion. This reverses the reaction polarity. \\
\hline
5 & \textbf{Type:} acid\_base\_proton\_transfer \newline
    \textbf{Intermediate:} \texttt{C=CC1CC1} &
    Produces the correct product but through a nonphysical sequence of steps. The mechanism lacks heterolytic cleavage and proper diazonium formation. \\
\bottomrule
\end{tabular}
}
\caption{Model-generated mechanism for Case Study 4.}
\label{tab:bamford_case}
\end{table}

\paragraph{Expert Analysis}  
In the gold-standard mechanism, the key step is the heterolytic cleavage of the diazonium intermediate to form a carbocation, followed by base-promoted proton elimination. The model, however, omits the cleavage and instead chains multiple proton-transfer and elimination operations. This yields the correct alkene but through a chemically implausible pathway, illustrating that the model captures surface-level reactivity patterns without understanding charge redistribution or intermediate stability.

\newpage

\subsection{ICL Case Study}
\label{app:case-study-icl}

\textbf{Reactants:} \texttt{Nc1ccc(C(=O)C(=O)c2ccccc2)cc1} \\
\textbf{Products:} \texttt{Nc1ccc(C(O)(C(=O)O)c2ccccc2)cc1} \\
\textbf{Conditions:} NaOH, H$_2$O

\begin{table}[htbp]
\centering
\small
\resizebox{\textwidth}{!}{
\begin{tabular}{p{0.05\textwidth} p{0.45\textwidth} p{0.45\textwidth}}
\toprule
\textbf{Step} & \textbf{Without ICL} & \textbf{With ICL (Aligned with Gold)} \\
\midrule
1 & \textbf{Type:} nucleophilic\_addition \newline
\textbf{Intermediate:} \texttt{Nc1ccc(C(=O)C([O-])(O)c2ccccc2)cc1} &
\textbf{Type:} nucleophilic\_addition \newline
\textbf{Intermediate:} \texttt{Nc1ccc(C(=O)C([O-])(O)c2ccccc2)cc1} \\
\hline
2 & \textbf{Type:} 1,2-shift \newline
\textbf{Intermediate:} \texttt{Nc1ccc(C(O)(C(=O)[O-])c2ccccc2)cc1} &
\textbf{Type:} 1,2-shift \newline
\textbf{Intermediate:} \texttt{O=C(O)C([O-])(c1ccccc1)c1ccc(N)cc1} \\
\hline
3 & \textbf{Type:} acid\_base\_proton\_transfer \newline
\textbf{Intermediate:} \texttt{Nc1ccc(C(O)(C(=O)O)c2ccccc2)cc1} &
\textbf{Type:} acid\_base\_proton\_transfer \newline
\textbf{Intermediate:} \texttt{Nc1ccc(C(O)(C(=O)[O-])c2ccccc2)cc1} \\
\hline
4 & -- (missing) &
\textbf{Type:} acid\_base\_proton\_transfer \newline
\textbf{Intermediate:} \texttt{Nc1ccc(C(O)(C(=O)O)c2ccccc2)cc1} \\
\bottomrule
\end{tabular}
}
\caption{A case exemplifies how ICL improves (PT denotes proton transfer). Without ICL, the model omits the critical rearrangement step and terminates early after three steps.}
\end{table}

\textbf{Expert Analysis.}  
Without ICL, the model halts prematurely and fails to include the second proton-transfer step, leaving the mechanism incomplete. With ICL, the model recovers the correct rearrangement pathway and produces all four steps, mirroring the gold-standard mechanism. This example demonstrates that exemplar-guided prompting enhances causal continuity and encourages models to reconstruct multi-step reasoning chains consistent with chemical logic.
\newpage

\section{Results}

This section presents the performance outcomes of various large language models on \textbf{oMeBench} under different reasoning and training configurations. We evaluate models across multiple experimental settings, including \emph{chain-of-thought prompting} (CoT), \emph{in-context learning} (ICL), and \emph{fine-tuning}, to examine how these paradigms influence mechanistic reasoning ability. Each subsection reports quantitative metrics covering validity, linguistic correctness, and both total and partial structural alignment, organized by difficulty level or data domain.

\subsection{Main results}

\begin{figure*}[!htb]
    \centering
    \includegraphics[width=\linewidth]{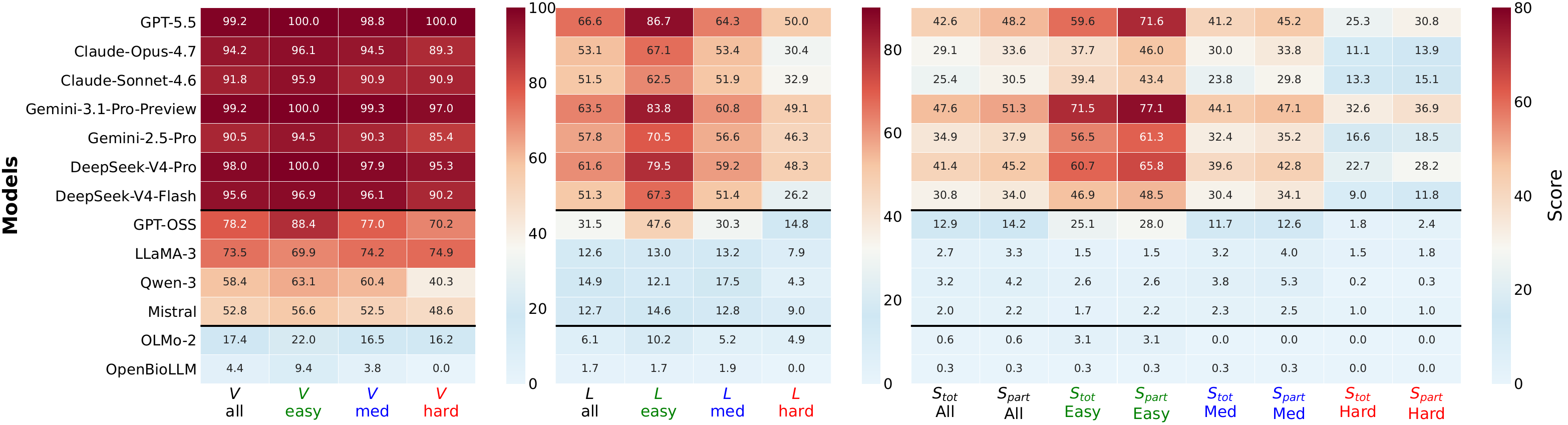}
    \caption{Performance of LLMs on oMeBench across difficulty levels. Frontier models outperform others but all degrade on harder reactions, while chemistry-specific models lag despite domain pretraining}
    \vspace{-3mm}
    \label{fig:heatmap}
\end{figure*}

Tab.~\ref{tab:results_full} and Fig.~\ref{fig:heatmap} shows the full result of our evaluation.

\begin{table*}[!htb]
\centering

\small
\resizebox{\textwidth}{!}{%
\begin{tabular}{@{}l r rrrr rrrr rrrr rrrr@{}}
\toprule
\multirow{2}{*}{\textbf{Model}} & \multirow{2}{*}{\textbf{Size}} & \multicolumn{4}{c}{\textbf{Overall}} & \multicolumn{4}{c}{\textbf{Easy}} & \multicolumn{4}{c}{\textbf{Medium}} & \multicolumn{4}{c}{\textbf{Hard}} \\
\cmidrule(lr){3-6} \cmidrule(lr){7-10} \cmidrule(lr){11-14} \cmidrule(lr){15-18}
 & & \textbf{V} & \textbf{L} & \textbf{$S_{tot}$} & \textbf{$S_{part}$} & \textbf{V} & \textbf{L} & \textbf{$S_{tot}$} & \textbf{$S_{part}$} & \textbf{V} & \textbf{L} & \textbf{$S_{tot}$} & \textbf{$S_{part}$} & \textbf{V} & \textbf{L} & \textbf{$S_{tot}$} & \textbf{$S_{part}$} \\
\midrule
\multicolumn{18}{c}{\textbf{Proprietary and Frontier Models}} \\
\midrule
GPT-5.5 & - & \textbf{99.2} & \textbf{66.6} & \underline{42.6} & \underline{48.2} & \textbf{100.0} & \textbf{86.7} & 59.6 & \underline{71.6} & \underline{98.8} & \textbf{64.3} & \underline{41.2} & \underline{45.2} & \textbf{100.0} & \textbf{50.0} & \underline{25.3} & \underline{30.8} \\
Claude-Opus-4.7 & - & 94.2 & 53.1 & 29.1 & 33.6 & 96.1 & 67.1 & 37.7 & 46.0 & 94.5 & 53.4 & 30.0 & 33.8 & 89.3 & 30.4 & 11.1 & 13.9 \\
Claude-Sonnet-4.6 & - & 91.8 & 51.5 & 25.4 & 30.5 & 95.9 & 62.5 & 39.4 & 43.4 & 90.9 & 51.9 & 23.8 & 29.8 & 90.9 & 32.9 & 13.3 & 15.1 \\
Gemini-3.1-Pro-Preview & - & \textbf{99.2} & \underline{63.5} & \textbf{47.6} & \textbf{51.3} & \textbf{100.0} & \underline{83.8} & \textbf{71.5} & \textbf{77.1} & \textbf{99.3} & \underline{60.8} & \textbf{44.1} & \textbf{47.1} & \underline{97.0} & \underline{49.1} & \textbf{32.6} & \textbf{36.9} \\
Gemini-2.5-Pro    & -   & 90.5 & 57.8 & 34.9 & 37.9 & 94.5 & 70.5 & 56.5 & 61.3 & 90.3 & 56.6 & 32.4 & 35.2 & 85.4 & 46.3 & 16.6 & 18.5 \\
DeepSeek-V4-Pro & - & \underline{98.0} & 61.6 & 41.4 & 45.2 & \textbf{100.0} & 79.5 & \underline{60.7} & 65.8 & 97.9 & 59.2 & 39.6 & 42.8 & 95.3 & 48.3 & 22.7 & 28.2 \\
DeepSeek-V4-Flash & - & 95.6 & 51.3 & 30.8 & 34.0 & \underline{96.9} & 67.3 & 46.9 & 48.5 & 96.1 & 51.4 & 30.4 & 34.1 & 90.2 & 26.2 & 9.0 & 11.8 \\
GPT-5 & - & 90.4 & 51.9 & 23.6 & 29.1 & 92.4 & 69.9 & 42.5 & 51.8 & 90.0 & 50.7 & 21.6 & 26.5 & 90.2 & 31.5 & 7.2 & 10.0 \\
GPT-4o & -   & 83.7 & 20.3 & 3.9 & 5.0 & 82.6 & 26.0 & 8.3 & 10.1 & 83.8 & 20.6 & 3.2 & 4.2 & 85.1 & 9.7 & 1.6 & 1.6 \\
o3 & -   & 90.8 & 47.4 & 24.0 & 28.1 & 94.6 & 64.5 & 36.8 & 47.3 & 90.6 & 45.6 & 23.3 & 26.3 & 86.4 & 32.5 & 8.6 & 9.7 \\
Claude-Sonnet-4   & - & 91.0 & 37.3 & 14.9 & 17.9 & 93.0 & 54.3 & 28.9 & 33.9 & 90.9 & 35.8 & 12.7 & 15.6 & 88.7 & 20.7 & 7.5 & 7.9 \\
DeepSeek-R1* & 685B & 87.6 & 45.4 & 22.5 & 25.0 & 88.4 & 62.3 & 39.0 & 41.9 & 87.9 & 45.1 & 20.6 & 23.3 & 82.9 & 21.3 & 8.3 & 9.7 \\
\midrule
\multicolumn{18}{c}{\textbf{Open Source Models}} \\
\midrule
GPT-OSS & 20B & 78.2 & 31.5 & 12.9 & 14.2 & 88.4 & 47.6 & 25.1 & 28.0 & 77.0 & 30.3 & 11.7 & 12.6 & 70.2 & 14.8 & 1.8 & 2.4 \\
LLaMA-3 & 8B & 73.5 & 12.6 & 2.7 & 3.3 & 69.9 & 13.0 & 1.5 & 1.5 & 74.2 & 13.2 & 3.2 & 4.0 & 74.9 & 7.9 & 1.5 & 1.8 \\
Qwen-3 & 4B & 58.4 & 14.9 & 3.2 & 4.2 & 63.1 & 12.1 & 2.6 & 2.6 & 60.4 & 17.5 & 3.8 & 5.3 & 40.3 & 4.3 & 0.2 & 0.3 \\
Mistral & 7B & 52.8 & 12.7 & 2.0 & 2.2 & 56.6 & 14.6 & 1.7 &  2.2 & 52.5 & 12.8 & 2.3 & 2.5 & 48.6 & 9.0 & 1.0 & 1.0 \\
Phi-4 & 7B & 0.3 & 0.0 & 0.0 & 0.0 & 0.0 & 0.0 & 0.0 & 0.0 & 0.3 & 0.0 & 0.0 & 0.0 & 0.0 & 0.0 & 0.0 & 0.0 \\
\midrule
\multicolumn{18}{c}{\textbf{Domain-Specific Models}} \\
\midrule
OLMo-2 & 13B & 17.4 & 6.1 & 0.6 & 0.6 & 22.0 & 10.2 & 3.1 & 3.1 & 16.5 & 5.2 & 0.0 & 0.0 & 16.2 & 4.9 & 0.0 & 0.0 \\
OpenBioLLM & 8B & 4.4 & 1.7 & 0.3 & 0.3 & 9.4 & 1.7 & 0.3 & 0.3 & 3.8 & 1.9 & 0.3 & 0.3 & 0.0 & 0.0 & 0.0 & 0.0 \\
ChemDFM & 8B & 0.0 & 0.7 & 0.0 & 0.0 & 0.0 & 0.0 & 0.0 & 0.0 & 0.7 & 0.0 & 0.0 & 0.0 & 0.0 & 0.0 & 0.0 & 0.0 \\
BioT5+ & 770M & 0.0 & 0.0 & 0.0 & 0.0 & 0.0 & 0.0 & 0.0 & 0.0 & 0.0 & 0.0 & 0.0 & 0.0 & 0.0 & 0.0 & 0.0 & 0.0 \\
\bottomrule
\end{tabular}
}
\caption{Model Performance on oMeBench Across Difficulty Levels}
\label{tab:results_full}
\vspace{-5mm}
\end{table*}

\subsection{CoT results}
\label{app:cot_results}

This subsection summarizes model performance under the one-shot \emph{chain-of-thought} setting, where models are prompted to explicitly reason through each mechanistic step before predicting the next intermediate or transformation.

\begin{table*}[!h]
\centering
\small
\resizebox{\textwidth}{!}{%
\begin{tabular}{@{}l r rrrr rrrr rrrr rrrr@{}}
\toprule
\multirow{2}{*}{\textbf{Model}} & \multirow{2}{*}{\textbf{Size}} 
& \multicolumn{4}{c}{\textbf{Overall}} 
& \multicolumn{4}{c}{\textbf{Easy}} 
& \multicolumn{4}{c}{\textbf{Medium}} 
& \multicolumn{4}{c}{\textbf{Hard}} \\
\cmidrule(lr){3-6} \cmidrule(lr){7-10} \cmidrule(lr){11-14} \cmidrule(lr){15-18}
& & \textbf{V} & \textbf{L} & \textbf{$S_{tot}$} & \textbf{$S_{part}$}
& \textbf{V} & \textbf{L} & \textbf{$S_{tot}$} & \textbf{$S_{part}$}
& \textbf{V} & \textbf{L} & \textbf{$S_{tot}$} & \textbf{$S_{part}$}
& \textbf{V} & \textbf{L} & \textbf{$S_{tot}$} & \textbf{$S_{part}$} \\
\midrule
\multicolumn{18}{c}{\textbf{Open Source Models (One-shot CoT Setting)}} \\
\midrule
Mistral & 7B & 55.2 & 10.1 & 3.0 & 3.5 & 53.8 & 20.8 & 5.1 & 5.5 & 56.0 & 7.3 & 2.7 & 3.3 & 53.0 & 10.4 & 1.3 & 1.9 \\
Qwen-3 & 4B & 49.9 & 12.6 & 2.6 & 3.2 & 54.3 & 14.7 & 1.5 & 2.6 & 49.0 & 13.2 & 3.1 & 3.8 & 48.9 & 6.0 & 0.9 & 1.1 \\
OLMo-2 & 13B & 18.7 & 6.9 & 1.1 & 1.1 & 13.5 & 9.2 & 2.9 & 2.9 & 21.1 & 6.6 & 0.8 & 0.9 & 12.1 & 5.8 & 0.0 & 0.0 \\
\bottomrule
\end{tabular}
}
\caption{Performance of Open-Source Models that work under One-shot Chain-of-Thought (CoT) Setting across Difficulty Levels.}
\label{tab:oneshot_cot_results}
\end{table*}

\subsection{In-context learning (ICL) results}
\label{app:icl_results}

This subsection reports results from the \emph{in-context learning} setting, where models are provided with several exemplars of annotated reaction mechanisms before generating predictions for new queries.

\begin{table*}[!h]
\centering
\small

\resizebox{\textwidth}{!}{%
\begin{tabular}{@{}l r rrrr rrrr rrrr rrrr@{}}
\toprule
\multirow{2}{*}{\textbf{Model}} & \multirow{2}{*}{\textbf{Size}} & \multicolumn{4}{c}{\textbf{Overall}} & \multicolumn{4}{c}{\textbf{Easy}} & \multicolumn{4}{c}{\textbf{Medium}} & \multicolumn{4}{c}{\textbf{Hard}} \\
\cmidrule(lr){3-6} \cmidrule(lr){7-10} \cmidrule(lr){11-14} \cmidrule(lr){15-18}
& & \textbf{V} & \textbf{L} & \textbf{$S_{tot}$} & \textbf{$S_{part}$} & \textbf{V} & \textbf{L} & \textbf{$S_{tot}$} & \textbf{$S_{part}$} & \textbf{V} & \textbf{L} & \textbf{$S_{tot}$} & \textbf{$S_{part}$} & \textbf{V} & \textbf{L} & \textbf{$S_{tot}$} & \textbf{$S_{part}$} \\
\midrule
\multicolumn{18}{c}{\textbf{Hypothetical and SOTA Frontier Models}} \\
\midrule
GPT-5.5 & - & \perf{99.9}{+0.7} & \perf{81.7}{+15.0} & \perf{65.0}{+22.4} & \perf{67.7}{+19.4} & \perf{100.0}{+0.0} & \perf{87.3}{+0.6} & \perf{76.5}{+17.0} & \perf{77.3}{+5.7} & \perf{99.9}{+1.0} & \perf{81.8}{+17.5} & \perf{63.8}{+22.6} & \perf{67.1}{+21.9} & \perf{100.0}{+0.0} & \perf{72.1}{+22.0} & \perf{55.0}{+29.7} & \perf{56.4}{+25.6} \\
Claude-Opus-4.7 & - & \perf{97.2}{+3.0} & \perf{77.7}{+24.6} & \perf{55.1}{+26.0} & \perf{58.6}{+24.9} & \perf{99.5}{+3.4} & \perf{81.3}{+14.2} & \perf{61.6}{+23.9} & \perf{63.2}{+17.2} & \perf{97.2}{+2.7} & \perf{77.4}{+24.0} & \perf{54.2}{+24.3} & \perf{58.3}{+24.6} & \perf{93.8}{+4.5} & \perf{74.3}{+43.9} & \perf{50.8}{+39.7} & \perf{53.0}{+39.1} \\
Claude-Sonnet-4.6 & - & \perf{94.3}{+2.5} & \perf{77.6}{+26.1} & \perf{55.1}{+29.7} & \perf{57.8}{+27.3} & \perf{99.1}{+3.1} & \perf{89.1}{+26.6} & \perf{70.2}{+30.8} & \perf{70.7}{+27.3} & \perf{93.8}{+2.9} & \perf{76.2}{+24.3} & \perf{53.6}{+29.7} & \perf{56.4}{+26.6} & \perf{90.3}{-0.5} & \perf{68.9}{+36.0} & \perf{41.3}{+28.1} & \perf{47.0}{+31.9} \\
Gemini-Pro-3.1-Preview & - & \perf{99.7}{+0.6} & \perf{83.1}{+19.5} & \perf{69.9}{+22.3} & \perf{71.9}{+20.6} & \perf{100.0}{+0.0} & \perf{89.8}{+6.0} & \perf{81.8}{+10.3} & \perf{85.2}{+8.1} & \perf{99.7}{+0.4} & \perf{82.7}{+21.9} & \perf{68.3}{+24.2} & \perf{69.8}{+22.7} & \perf{99.4}{+2.4} & \perf{74.9}{+25.8} & \perf{61.9}{+29.3} & \perf{63.7}{+26.8} \\
Gemini-Pro-2.5 & - & \perf{95.2}{+4.7} & \perf{73.8}{+16.0} & \perf{54.4}{+19.5} & \perf{56.9}{+19.0} & \perf{97.5}{+3.0} & \perf{79.9}{+9.4} & \perf{65.7}{+9.2} & \perf{69.4}{+8.1} & \perf{94.9}{+4.6} & \perf{74.8}{+18.2} & \perf{54.6}{+22.2} & \perf{56.8}{+21.6} & \perf{93.1}{+7.7} & \perf{58.4}{+12.1} & \perf{35.5}{+18.9} & \perf{38.4}{+19.9} \\
DeepSeek-V4-Pro & - & \perf{97.6}{-0.4} & \perf{80.6}{+19.0} & \perf{62.6}{+21.2} & \perf{64.8}{+19.6} & \perf{99.1}{-0.9} & \perf{87.5}{+8.0} & \perf{79.8}{+19.1} & \perf{80.3}{+14.5} & \perf{97.3}{-0.6} & \perf{80.2}{+21.0} & \perf{60.6}{+21.1} & \perf{62.6}{+19.8} & \perf{96.7}{+1.5} & \perf{72.0}{+23.7} & \perf{48.1}{+25.4} & \perf{54.7}{+26.5} \\
DeepSeek-V4-Flash & - & \perf{96.1}{+0.5} & \perf{76.3}{+25.0} & \perf{57.2}{+26.4} & \perf{59.9}{+25.8} & \perf{100.0}{+3.1} & \perf{86.8}{+19.4} & \perf{71.6}{+24.7} & \perf{74.4}{+25.9} & \perf{95.2}{-0.9} & \perf{74.5}{+23.1} & \perf{54.5}{+24.1} & \perf{57.5}{+23.5} & \perf{95.3}{+5.1} & \perf{71.1}{+44.9} & \perf{51.5}{+42.5} & \perf{52.0}{+40.2} \\
\midrule
\multicolumn{18}{c}{\textbf{Open Source Models}} \\
\midrule
GPT-OSS & 20B & \perf{79.3}{+1.1} & \perf{43.1}{+11.6} & \perf{23.1}{+10.2} & \perf{24.0}{+9.8} & \perf{79.0}{-9.4} & \perf{55.6}{+8.0} & \perf{34.7}{+9.6} & \perf{34.9}{+6.9} & \perf{80.0}{+3.0} & \perf{42.9}{+12.6} & \perf{22.6}{+10.9} & \perf{23.6}{+11.0} & \perf{75.6}{+5.4} & \perf{25.0}{+10.2} & \perf{8.6}{+6.8} & \perf{9.6}{+7.2} \\
LLaMA-3 & 8B & \perf{77.2}{+3.7} & \perf{44.5}{+31.9} & \perf{19.0}{+16.3} & \perf{20.8}{+17.5} & \perf{77.2}{+7.3} & \perf{40.1}{+27.1} & \perf{16.4}{+14.9} & \perf{20.0}{+18.5} & \perf{76.4}{+2.2} & \perf{46.0}{+32.8} & \perf{20.3}{+17.1} & \perf{21.8}{+17.8} & \perf{73.7}{-1.2} & \perf{41.7}{+33.8} & \perf{15.0}{+13.5} & \perf{15.9}{+14.1} \\
Qwen-3 & 4B & \perf{75.6}{+17.2} & \perf{50.1}{+35.2} & \perf{24.6}{+21.4} & \perf{27.0}{+22.8} & \perf{78.8}{+15.7} & \perf{53.1}{+41.0} & \perf{30.5}{+27.9} & \perf{34.3}{+31.7} & \perf{75.7}{+15.3} & \perf{49.8}{+32.3} & \perf{24.9}{+21.1} & \perf{27.0}{+21.7} & \perf{67.0}{+26.7} & \perf{47.2}{+42.9} & \perf{14.0}{+13.8} & \perf{15.7}{+15.4} \\
Mistral & 7B & \perf{63.7}{+10.9} & \perf{38.9}{+26.2} & \perf{16.2}{+14.2} & \perf{17.9}{+15.7} & \perf{68.5}{+11.9} & \perf{42.4}{+27.8} & \perf{20.1}{+18.4} & \perf{23.4}{+21.2} & \perf{63.0}{+10.5} & \perf{39.6}{+26.8} & \perf{16.3}{+14.0} & \perf{17.6}{+15.1} & \perf{64.0}{+15.4} & \perf{29.6}{+20.6} & \perf{10.0}{+9.0} & \perf{11.2}{+10.2} \\
\midrule
\multicolumn{18}{c}{\textbf{Chemistry-specified Models}} \\
\midrule
OLMo-2 & 13B & \perf{53.5}{+36.1} & \perf{31.1}{+25.0} & \perf{10.9}{+10.3} & \perf{12.3}{+11.7} & \perf{54.9}{+32.9} & \perf{33.8}{+23.6} & \perf{12.5}{+9.4} & \perf{14.6}{+11.5} & \perf{53.9}{+37.4} & \perf{32.1}{+26.9} & \perf{10.9}{+10.9} & \perf{12.1}{+12.1} & \perf{48.6}{+32.4} & \perf{21.0}{+16.1} & \perf{8.6}{+8.6} & \perf{9.7}{+9.7} \\
\bottomrule
\end{tabular}%
}
\\[2ex]
\parbox{\textwidth}{%
\textit{\footnotesize{Score Metrics: \textbf{V} (Validity), \textbf{L} (Linguistic Correctness), \textbf{$S_{tot}$} (Total Structural Match), \textbf{$S_{part}$} (Partial Structural Match). (Assumed definitions)}} \\
\textit{\footnotesize{The second line in each cell shows the change (+/-) from the baseline performance.}}
}
\caption{Model Performance (ICL) with Change from Baseline}

\label{tab:icl_model_performance}
\end{table*}

\newpage
\subsection{ICL Retrieval Ablation}\label{app:icl-ablation}
To assess whether similarity-based retrieval is truly 
necessary, we compare ICL-DRFP with an ICL-Random baseline 
using randomly sampled Silver reactions as demonstrations. 
Table~\ref{tab:icl-ablation} reports the results, where 
B denotes the zero-shot baseline, $\Delta$R = ICL-Random $-$ Base, 
and $\Delta$D = ICL-DRFP $-$ ICL-Random.

\begin{table}[H]
\centering
\resizebox{\linewidth}{!}{%
\begin{tabular}{lcccc}
\toprule
\textbf{Model} & \textbf{V (B/$\Delta$R/$\Delta$D)} & \textbf{L (B/$\Delta$R/$\Delta$D)} & $\boldsymbol{S_{tot}}$ \textbf{(B/$\Delta$R/$\Delta$D)} & $\boldsymbol{S_{part}}$ \textbf{(B/$\Delta$R/$\Delta$D)} \\
\midrule
GPT-5       & 90.4/+0.1/+4.1   & 51.9/$-$1.1/+15.5  & 23.6/$-$0.6/+23.9  & 29.1/$-$0.3/+21.8 \\
GPT-4o      & 83.7/$-$0.4/$-$5.0 & 20.3/+1.3/+22.4   & 3.9/+1.2/+18.1    & 5.0/+1.1/+19.9   \\
Gemini-2.5  & 90.5/+3.1/+1.6   & 57.8/+3.3/+12.7   & 34.9/+1.7/+17.8   & 37.9/+3.4/+15.6  \\
DeepSeek-R1 & 87.6/+1.3/+1.7   & 45.4/+4.8/+18.5   & 22.5/+2.9/+20.2   & 25.0/+3.4/+19.8  \\
\midrule
LLaMA-3     & 73.5/$-$1.1/+4.8  & 12.6/+1.6/+30.3   & 2.7/+0.9/+15.4    & 3.3/+0.6/+16.9   \\
Qwen-3      & 58.4/+10.9/+6.3  & 14.9/+2.5/+32.7   & 3.2/+0.7/+20.7    & 4.2/+1.3/+21.5   \\
OLMo-2      & 17.4/+24.5/+11.6 & 6.1/+7.5/+17.5    & 0.6/+1.9/+8.4     & 0.6/+2.6/+9.1    \\
Mistral     & 52.8/$-$1.2/+12.1 & 12.7/+0.7/+25.5   & 2.0/+1.0/+13.2    & 2.2/+1.1/+14.6   \\
\bottomrule
\end{tabular}%
}
\caption{Ablation study of ICL retrieval strategies. ICL-Random 
uses randomly sampled Silver reactions; ICL-DRFP retrieves 
examples by DRFP similarity.}
\label{tab:icl-ablation}
\end{table}

\noindent\textbf{Key Observations.} 
(1)~ICL-Random yields only marginal and unstable gains. Random demonstrations help models follow the output format but often introduce irrelevant chemical contexts, leading to noisy or even negative effects.
(2)~ICL-DRFP yields substantial and consistent improvements. High-similarity examples reflect mechanistically relevant patterns (reactive-center alignment, electron-flow motifs, plausible intermediates), acting as soft mechanistic priors and improving structural coherence. 

\newpage

\subsection{Fine-tuning results}
\label{app:finetune_results}

This subsection presents results from \emph{fine-tuned} variants of selected models, comparing their pre- and post-tuning performance under both simple and in-context prompting conditions.

\begin{table*}[!h]
\centering
\resizebox{\textwidth}{!}{%
\begin{tabular}{@{}llcccccccccccc@{}}
\toprule
\multirow{2}{*}{\textbf{Model}} & \multirow{2}{*}{\textbf{Size}} 
& \multicolumn{4}{c}{\textbf{Overall}} 
& \multicolumn{4}{c}{\textbf{In-Domain}} 
& \multicolumn{4}{c}{\textbf{Out-of-Domain}} \\
\cmidrule(lr){3-6} \cmidrule(lr){7-10} \cmidrule(lr){11-14}
& & \textbf{V} & \textbf{L} & \textbf{$S_{tot}$} & \textbf{$S_{part}$} 
  & \textbf{V} & \textbf{L} & \textbf{$S_{tot}$} & \textbf{$S_{part}$} 
  & \textbf{V} & \textbf{L} & \textbf{$S_{tot}$} & \textbf{$S_{part}$} \\
\midrule
\multicolumn{14}{c}{\textbf{Before Fine-tuning (Simple Prompt)}} \\
\midrule
Qwen-3          & 4B & 58.5 & 15.0 & 3.2 & 4.3 & 59.8 & 16.2 & 3.0 & 4.3 & 54.7 & 11.4 & 3.8 & 4.1 \\
LLaMA-3         & 8B & 73.5 & 12.6 & 2.7 & 3.3 & 76.9 & 12.7 & 2.6 & 2.9 & 63.6 & 12.0 & 2.9 & 4.3 \\
Mistral         & 7B & 52.8 & 12.7 & 2.0 & 2.2 & 54.0 & 13.0 & 2.1 & 2.3 & 49.0 & 11.9 & 1.8 & 2.1 \\
OLMo-2          & 13B & 17.4 & 6.1 & 0.6 & 0.6 & 19.4 & 6.4 & 0.8 & 0.8 & 11.6 & 5.2 & 0.0 & 0.0 \\
\midrule
\multicolumn{14}{c}{\textbf{After Fine-tuning (Simple Prompt)}} \\
\midrule
Qwen-3-SFT      & 4B & \perf{85.2}{+26.7} & \perf{38.7}{+23.7} & \perf{18.2}{+15.0} & \perf{20.6}{+16.3} & \perf{86.8}{+27.0} & \perf{43.2}{+27.0} & \perf{20.9}{+17.9} & \perf{23.5}{+19.2} & \perf{80.1}{+25.4} & \perf{24.7}{+13.3} & \perf{9.7}{+5.9} & \perf{11.5}{+7.4} \\
Qwen-3-CoT-SFT  & 4B & \perf{82.6}{+24.1} & \perf{37.5}{+22.5} & \perf{15.5}{+12.3} & \perf{17.4}{+13.1} & \perf{83.3}{+23.5} & \perf{42.0}{+25.8} & \perf{18.9}{+15.9} & \perf{20.6}{+16.3} & \perf{80.6}{+25.9} & \perf{24.1}{+12.7} & \perf{5.3}{+1.5} & \perf{7.8}{+3.7} \\
LLaMA-3-LoRA    & 8B & \perf{65.0}{-8.5} & \perf{19.6}{+7.0} & \perf{3.9}{+1.2} & \perf{4.8}{+1.5} & \perf{64.7}{-12.2} & \perf{21.7}{+9.0} & \perf{4.4}{+1.8} & \perf{5.3}{+2.4} & \perf{66.0}{+2.4} & \perf{13.5}{+1.5} & \perf{2.3}{-0.6} & \perf{3.4}{-0.9} \\
Mistral-LoRA    & 7B & \perf{77.8}{+25.0} & \perf{19.4}{+6.7} & \perf{6.5}{+4.5} & \perf{7.6}{+5.4} & \perf{79.3}{+25.3} & \perf{21.4}{+8.4} & \perf{7.9}{+5.8} & \perf{9.2}{+6.9} & \perf{73.5}{+24.5} & \perf{13.3}{+1.4} & \perf{2.2}{+0.4} & \perf{2.7}{+0.6} \\
OLMo-2-LoRA     & 13B & \perf{75.7}{+58.3} & \perf{17.8}{+11.7} & \perf{6.3}{+5.7} & \perf{7.4}{+6.8} & \perf{76.9}{+59.5} & \perf{18.7}{+12.3} & \perf{7.4}{+6.8} & \perf{8.3}{+7.7} & \perf{72.0}{+60.4} & \perf{15.0}{+9.8} & \perf{2.7}{+2.7} & \perf{4.7}{+4.7} \\
\midrule
\multicolumn{14}{c}{\textbf{Before Fine-tuning (ICL Baseline)}} \\
\midrule
Qwen-3          & 4B & 75.2 & 50.1 & 24.6 & 27.0 & 78.4 & 60.3 & 30.8 & 32.9 & 65.8 & 19.4 & 6.0 & 9.2 \\
LLaMA-3         & 8B & 76.2 & 44.5 & 19.0 & 20.8 & 78.1 & 52.6 & 23.7 & 25.6 & 70.5 & 20.2 & 4.8 & 6.3 \\
Mistral         & 7B & 64.1 & 38.9 & 16.2 & 17.9 & 67.6 & 48.1 & 20.7 & 22.9 & 53.7 & 11.3 & 2.9 & 2.9 \\
OLMo-2          & 13B & 53.5 & 31.1 & 10.9 & 12.3 & 57.2 & 36.8 & 13.3 & 15.0 & 42.3 & 14.0 & 3.9 & 4.1 \\
\midrule
\multicolumn{14}{c}{\textbf{After Fine-tuning (with ICL)}} \\
\midrule
Qwen-3-SFT      & 4B & \perf{84.4}{+9.2} & \perf{53.6}{+3.5} & \perf{27.4}{+2.8} & \perf{30.6}{+3.6} & \perf{84.7}{+6.3} & \perf{63.3}{+3.0} & \perf{33.8}{+3.0} & \perf{36.5}{+3.6} & \perf{83.7}{+17.9} & \perf{24.4}{+5.0} & \perf{8.1}{+2.1} & \perf{12.9}{+3.7} \\
Qwen-3-CoT-SFT  & 4B & \perf{85.6}{+10.4} & \perf{55.1}{+5.0} & \perf{29.2}{+4.6} & \perf{31.0}{+4.0} & \perf{86.5}{+8.1} & \perf{65.5}{+5.2} & \perf{36.3}{+5.5} & \perf{38.0}{+5.1} & \perf{83.0}{+17.2} & \perf{23.8}{+4.4} & \perf{7.8}{+1.8} & \perf{9.9}{+0.7} \\
LLaMA-3-LoRA    & 8B & \perf{83.2}{+7.0} & \perf{51.5}{+7.0} & \perf{26.8}{+7.8} & \perf{28.6}{+7.8} & \perf{86.7}{+8.6} & \perf{61.7}{+9.1} & \perf{33.2}{+9.5} & \perf{35.5}{+9.9} & \perf{73.0}{+2.5} & \perf{21.2}{+1.0} & \perf{7.7}{+2.9} & \perf{7.9}{+1.6} \\
Mistral-LoRA    & 7B & \perf{82.5}{+18.4} & \perf{45.0}{+6.1} & \perf{22.2}{+6.0} & \perf{24.8}{+6.9} & \perf{84.5}{+16.9} & \perf{55.1}{+7.0} & \perf{28.5}{+7.8} & \perf{31.9}{+9.0} & \perf{76.4}{+22.7} & \perf{15.0}{+3.7} & \perf{3.2}{+0.3} & \perf{3.4}{+0.5} \\
OLMo-2-LoRA     & 13B & \perf{79.7}{+26.2} & \perf{39.3}{+8.2} & \perf{21.3}{+10.4} & \perf{22.8}{+10.5} & \perf{82.4}{+25.2} & \perf{46.8}{+10.0} & \perf{25.6}{+12.3} & \perf{27.4}{+12.4} & \perf{71.6}{+29.3} & \perf{16.8}{+2.8} & \perf{8.2}{+4.3} & \perf{8.9}{+4.8} \\
\bottomrule
\multicolumn{14}{l}{\rule{0pt}{2.5ex}\textit{\footnotesize{Score Metrics: \textbf{V} (Validity), \textbf{L} (Linguistic Correctness), \textbf{$S_{tot}$} (Total Structural Match), \textbf{$S_{part}$} (Partial Structural Match).}}} \\
\end{tabular}%
}
\caption{Performance of fine-tuned models under one-shot simple prompt and three-shot ICL prompt}
\label{tab:fine-tune-optimized}
\end{table*}

\newpage
\twocolumn
\section{Fine-Tuning Configuration and Hyperparameters}
\label{sec:fine-tuning}

We fine-tune large language models (LLMs) for chemical reaction mechanism prediction using two complementary parameter-efficient strategies: full supervised fine-tuning (SFT) and Low-Rank Adaptation (LoRA).

\subsection{Dataset and Preprocessing}

Both configurations utilize the \texttt{alpaca\_chemical\_mechanism\_dataset}, formatted in the Alpaca instruction-following template~\citep{taori2023alpaca} with three fields: \textit{instruction} (task description), \textit{input} (reactants and products in SMILES format), and \textit{output} (predicted reaction mechanism). 

To accommodate the structured nature of chemical reactions and their detailed mechanistic descriptions, we set a maximum sequence length of $L_{\text{max}} = 6000$ tokens, substantially longer than typical instruction-following tasks. We employ 16 parallel workers for data preprocessing to efficiently handle the large-scale chemical dataset.

\subsection{Optimization Hyperparameters}

Table~\ref{tab:hyperparameters} summarizes the key hyperparameters for both fine-tuning strategies.

\paragraph{Learning Rate and Schedule:}
We adopt substantially different learning rates for the two strategies. For full SFT, we use a conservative learning rate of $\eta = 1 \times 10^{-5}$ to prevent catastrophic forgetting of the model's pre-trained chemical and general knowledge. In contrast, LoRA fine-tuning employs $\eta = 1 \times 10^{-4}$, an order of magnitude larger, as only the low-rank adaptation matrices are updated and thus require stronger gradient signals.

Both configurations use a cosine learning rate schedule with a warmup ratio of 0.1, where the learning rate linearly increases from 0 to $\eta$ during the first 10\% of training steps, then follows a cosine decay to 0. This schedule stabilizes early training and prevents aggressive weight updates near convergence.

\paragraph{Distributed Training and Mixed Precision}

Both configurations leverage DeepSpeed ZeRO-3~\cite{rajbhandari2020zero} for memory-efficient distributed training. 

\begin{table}[htb]
\centering
\resizebox{\linewidth}{!}{
\begin{tabular}{lcc}
\toprule
\textbf{Hyperparameter} & \textbf{Full SFT} & \textbf{LoRA} \\
\midrule
\multicolumn{3}{l}{\textit{Optimization}} \\
Learning rate ($\eta$) & $1.0 \times 10^{-5}$ & $1.0 \times 10^{-4}$ \\
LR scheduler & Cosine & Cosine \\
Warmup ratio & 0.1 & 0.1 \\
Number of epochs & 6 & 8 \\
\midrule
\multicolumn{3}{l}{\textit{Batch Configuration}} \\
Per-device batch size & 1 & 1 \\
Gradient accumulation steps & 16 & 8 \\
Effective batch size & 16 & 8 \\
\midrule
\multicolumn{3}{l}{\textit{LoRA-specific}} \\
LoRA rank ($r$) & --- & 8 \\
LoRA target modules & --- & all \\
\midrule
\multicolumn{3}{l}{\textit{System}} \\
Precision & BF16 & BF16 \\
DeepSpeed stage & ZeRO-3 & ZeRO-3 \\
Max sequence length & 6000 & 6000 \\
\bottomrule
\end{tabular}}
\caption{Fine-tuning hyperparameters for full SFT (Qwen3-4B) and LoRA (> 4B).}
\label{tab:hyperparameters}
\end{table}

\end{document}